\documentclass[a4paper,12pt]{article}

\usepackage[utf8]{inputenc}
\usepackage[linesnumbered,ruled]{algorithm2e}
\let\oldnl\nl
\newcommand{\nonl}{\renewcommand{\nl}{\let\nl\oldnl}}
\usepackage{amsmath,amsthm,amssymb}
\usepackage{multirow, rotating}
\usepackage{adjustbox}

\usepackage{tikz}
\usetikzlibrary{decorations.pathreplacing,calc,tikzmark}
\usepackage{multirow}
\usepackage{pdflscape}

\usepackage{pgfplots}
\pgfplotsset{compat=newest}
\pgfplotscreateplotcyclelist{onemaxruntimes}{%
thick,blue\\ 
thick,blue,dashed\\ 
thick,red\\ 
thick,red,dashed\\  
thick,green!70!black\\ 
thick,green!70!black,dashed\\ 
}
\pgfplotscreateplotcyclelist{loruntimes}{%
thick,blue\\ 
thick,blue,dashed\\ 
thick,red\\ 
thick,red,dashed\\  
thick,green!70!black\\ 
thick,green!70!black,dashed\\ 
thick,orange\\ 
thick,orange,dashed\\ 
}
\pgfplotscreateplotcyclelist{optlamruntimes}{%
thick,blue\\ 
thick,blue,dashed\\ 
thick,red\\ 
thick,red,dashed\\ 
thick,green!70!black\\ 
thick,green!70!black,dashed\\ 
}
\pgfplotscreateplotcyclelist{jumpruntimes}{%
thick,blue\\ 
thick,red\\ 
thick,blue,dashed\\ 
thick,red,dashed\\ 
thick,green!70!black,dotted\\  
}
\usepgfplotslibrary{patchplots}

\newcommand{\oea}{\mbox{$(1 + 1)$~EA}\xspace}
\newcommand{\oplea}{\mbox{$(1+\lambda)$~EA}\xspace}
\newcommand{\mpoea}{\mbox{$(\mu+1)$~EA}\xspace}

\newcommand{\ollga}{${(1 + (\lambda , \lambda))}$~GA\xspace}

\newcommand{\onemax}{\textsc{OneMax}\xspace}

\newcommand{\leadingones}{\textsc{Leading\-Ones}\xspace}

\newcommand{\jump}{\textsc{Jump}\xspace}

\newcommand{\R}{{\mathbb R}}

\DeclareMathOperator{\Bin}{Bin}

\DeclareMathOperator*{\argmax}{arg\,max}

\begin{document}

\title{Larger Offspring Populations Help the $(1 + (\lambda, \lambda))$ Genetic Algorithm to Overcome the Noise}

\author{Alexandra Ivanova \\
        HSE University, Skoltech \\
  		Moscow, Russia \\
    \and
        Denis Antipov \\
		The University of Adelaide \\
        Adelaide, Australia \\
		\and
		Benjamin Doerr \\
		Laboratoire d'Informatique (LIX), \\
		CNRS, \'Ecole Polytechnique, \\ 
		Institut Polytechnique de Paris \\
		Palaiseau, France \\
} 

\maketitle
{\sloppy

\begin{abstract}
    Evolutionary algorithms are known to be robust to noise in the evaluation of the fitness. In particular, larger offspring population sizes often lead to strong robustness. We analyze to what extent the $(1+(\lambda,\lambda))$ genetic algorithm is robust to noise. This algorithm also works with larger offspring population sizes, but an intermediate selection step and a non-standard use of crossover as repair mechanism could render this algorithm less robust than, e.g., the simple $(1+\lambda)$ evolutionary algorithm. Our experimental analysis on several classic benchmark problems shows that this difficulty does not arise. Surprisingly, in many situations this algorithm is even more robust to noise than the $(1+\lambda)$~EA.
\end{abstract}


\section{Introduction}
\label{sec:intro}

Evolutionary algorithms (EAs) are general-purpose optimization heuristics. The facts that (i)~they use a large amount of independent randomness and (ii)~they do not exploit strongly the precise problem definition (they are so-called black-box optimizers) make it easy to believe that they are robust to all kinds of disturbances and, in fact, this has been observed multiple times~\cite{BianchiDGG09,JinB05}.

In this work, we concentrate on the most common stochastic disturbance, namely that the access to the objective function is prone to small stochastic errors. This is known as \emph{noisy function evaluations}. We also restrict ourselves to optimization in discrete search spaces, more precisely, to the search space $\Omega = \{0,1\}^n$ of bit strings of length~$n$, which is the most common representation in discrete evolutionary optimization. Since our focus is on gaining a solid understanding on how robust certain EAs are to noise, we restrict our analyses on classic benchmark problems. In this direction, most previous research results are mathematical runtime analyses, some however enriched with experimental investigations. We refer to the introduction of~\cite{Sudholt21} for a detailed account of the existing literature and describe here only the most relevant previous works.

The first mathematical runtime analysis of an EA in the presence of noise was conducted by Droste~\cite{Droste04}. It showed that the \oea can optimize the \onemax benchmark in polynomial time when noise appears with rate $O(\frac{\log n}{n})$. If the noise rate is asymptotically larger, superpolynomial runtimes result.

Gie{\ss}en and K\"otzing~\cite{GiessenK16} were the first to analyze the robustness of the simple population-based \mpoea and \oplea. For both, they were able to show much stronger runtime guarantees than for \oea when the population size was large. For example, they showed that for the one-bit noise with any rate $q \in (0, 1)$ the runtime of the \oplea on \onemax is $O(\frac{n^2\lambda}{q})$ if the population size $\lambda$ is at least $\max\{12/q, 24\}n\ln(n)$. Although this result does not work well for small noise rates $q = o(1)$, for all constant rates it shows that the polynomial runtime can be obtained with population size of order $\Omega(n\log(n))$. This is larger than the runtime of the \oplea on \onemax without noise shown in~\cite{JansenJW05} and~\cite{DoerrK15}, that is, $\Theta(n\log(n) + n\frac{\lambda \log^+\log^+(\lambda)}{\log^+(\lambda)})$, where $\log^+(x)$ stands for $\max\{1, \log(x)\}$. However, it is significantly better than the exponential runtime of the \oea for such large noise rates.

For the \leadingones benchmark, they could show a quadratic runtime for the \oea only for a noise rate $q \le 1/(6en^2)$, whereas for the \oplea with $72 \ln(n) \le \lambda = o(n)$ they showed this guarantee for all $q \le 0.028/n$. The very tight lower bounds proven in~\cite{Sudholt21} show that this discrepancy is real, namely, that the \oplea with moderate population size can indeed stand much higher noise levels than the \oea. Sudholt~\cite{Sudholt21} also greatly extended the upper bound of~\cite{GiessenK16} showing now, in particular, that a quadratic runtime is obtained when $\lambda \ge qn$ and $\lambda \ge 4.92 \ln(n)$. While no lower bounds were proven for the \oplea, the experiments in~\cite{Sudholt21} indicate that there is a clear threshold behavior so that the runtime explodes when the noise is too large, where ``too large'' depends on the population size~$\lambda$. We note that when there is no noise and $\lambda$ is at most polynomial in $n$, then the expected runtime of the \oplea on \leadingones is $\Theta(n^2 + \lambda n)$~\cite{JansenJW05}.

In this work, we continue the research direction started in~\cite{GiessenK16} and continued in~\cite{Sudholt21} by regarding how robust the \ollga is to noise. The \ollga is a genetic algorithm proposed first in~\cite{DoerrDE15}. Its main feature is that in each generation, it first generates $\lambda$ mutation offspring from the unique parent individual with a generally high mutation rate. It selects the best of these (``mutation winner'') and creates $\lambda$ new offspring via a biased uniform crossover between the parent and the mutation winner. Here the bias is such that bit values are more often taken from the parent. The best of these crossover offspring is the new parent individual unless the old parent is strictly better (in this case, the old parent is kept). This setup allows to use a higher mutation rate in the first phase, increasing the rate of exploration, since the biased crossover used in the second phase can act as repair mechanism and undo possible destruction from the more aggressive mutation. That this idea can indeed work out has been shown several times, most notably in the first works~\cite{DoerrDE15,DoerrD18}, where a small, but superconstant runtime gain on \onemax was shown (namely, the $O(\max\{\frac{n\log(n)}{\lambda}, \frac{n\lambda\log\log(\lambda)}{\log(\lambda)}\})$ runtime was shown, which with the right choice of $\lambda$ is by an $\Omega(\sqrt{\log(n)})$ factor smaller than the best possible runtime of the mutation-based EAs), in~\cite{BuzdalovD17} for (easy) random SAT instances, and in~\cite{AntipovDK22} with significant performance gains on jump functions. It is also worth mentioning that despite the \ollga relying on a strong correlation between the fitness and the distance to the optimum, it has the same $\Theta(n^2)$ runtime as most population-based algorithms on \leadingones, where this correlation is weak~\cite{AntipovDK19foga}. That the main working principle of the \ollga can also be exploited in multi-objective optimization, was shown in~\cite{DoerrHP22}.

What was not clear so far, and what is the focus of this work, is how robust the \ollga is to noise. The fact that in both phases of the algorithm $\lambda$ individuals are generated in parallel could mean that the algorithm inherits the robustness of the \oplea. On the other hand, the more complicated setup and in particular the intermediate selection step could also render the algorithm less robust. We note that there is no general rule that more complicated algorithms are less robust, but the fact that problem-specific algorithms, which usually are much more complex than simple evolutionary algorithms, are often not robust at all, points into this direction. We also feel that the analysis of the \ollga on easy random SAT instances~\cite{BuzdalovD17} suggests that this algorithm could be less robust. We note that the random SAT instances regarded there roughly give rise to fitness landscapes that resemble slightly disturbed \onemax instances. However, not the same results as in~\cite{DoerrDE15} could be shown, but certain adjustments to the algorithm where necessary to let it cope with the slightly more rugged fitness landscape of the random SAT instances.

Our main finding in this work is that these potential problems do not come true. We conduct an experimental analysis of the robustness question on the benchmarks \onemax, \leadingones, and \jump. These are the most common benchmarks in discrete evolutionary optimization, each with very different characteristics. They are known to show very different behaviors of the \ollga in the noise-free setting: Compared to simple mutation-based EAs such as the \oea or the \oplea, the \ollga has a small advantage on \onemax (with various ways to set the parameters~\cite{DoerrDE15,DoerrD18,AntipovBD22}), a huge advantage on \jump (when used with suitable parameters~\cite{AntipovDK22} or automated parameter choices~\cite{AntipovBD21gecco,AntipovD20ppsn}), and neither a significant advantage or disadvantage on \leadingones~\cite{AntipovDK19foga}.

On all three benchmarks, our experiments indicate that the \ollga has a good performance also in the presence of noise. Similar to the \oplea, roughly a logarithmic population size suffices. On \leadingones, both algorithms show a similar robustness, but for many settings, in particular, the easier ones, the \ollga suffers from the fact that each iteration is twice as costly (that is, requires $2\lambda$ fitness evaluations instead of~$\lambda$). This fits to the observation made already in~\cite{AntipovDK19foga} that the working principle of the \ollga is not effective on this problem. For the \onemax problem, also both algorithms show similar performance patterns, but in addition the \ollga keeps its advantage over the \oplea for logarithmic population sizes. On jump functions, the \ollga with the right parameters keeps its huge advantage (e.g., a speed-up by factor of $100$ for jump functions with problem size $2^7 = 128$ and gap size $k=3$)  over the \oplea for all noise intensities up to constant noise rates). A more detailed analysis on \onemax shows that already relatively small population sizes suffice to obtain robustness. For problems size $n = 1024$, the best results against constant-rate noise were obtained for population sizes between $5$ and $10$.

All these results indicate that the \ollga, despite its more complicated layout with two selection steps and a non-standard use of crossover, is highly robust against noise, even of high intensity, and this already from moderate population sizes on.

This works is organized as follows. In the next section, we introduce the benchmarks, the noise model, and the algorithms regarded in this work. In Sections~\ref{sec:onemax} to~\ref{sec:jump}, we describe our experimental results on the \onemax, \leadingones, and \jump benchmarks. A conclusion is presented in Section~\ref{sec:conclusion}.

\section{Problem Setting}
\label{sec:problem}
\subsection{Benchmark Problems}

In this paper we consider several pseudo-Boolean benchmark functions to investigate the robustness of the EAs to noise on different landscapes. All these functions are defined on a set of bit strings of length $n$ and return a real value. 

The first function we consider is the famous \onemax benchmark, which it returns the number of one-bits in its argument. More formally,
\begin{align*}
	\onemax(x) = \sum_{i = 1}^n x_i.
\end{align*}
\onemax has a clear fitness gradient towards optimum and thus it is often studied to understand how different algorithms perform on easy problems. While very simple, in fact, the simplest problem with unique global optimum in several respects~\cite{DoerrJW12algo,Sudholt13,Witt13}, this benchmark has nevertheless led to many important insights, e.g., on how optimal mutation rates could look like~\cite{Muhlenbein92}, how the selection pressure on non-elitist algorithms influences the runtime~\cite{Lehre10,Lehre11}, or that some natural EAs have enormous difficulties even with this simple benchmark~\cite{OlivetoW15,DoerrK20tec}.
Simple mutation-based EAs (with the parameters set appropriately) solve the \onemax problem in time $O(n \log n)$~\cite{Muhlenbein92,GarnierKS99,JansenJW05,Witt06,AntipovD21algo}, which is best-possible for a unary unbiased black-box algorithm~\cite{LehreW12}. Interestingly, also for many more complex algorithms such as ant-colony optimizers or estimation-of-distribution algorithms no better runtime than $O(n \log n)$ on \onemax could be shown~\cite{NeumannSW09,SudholtW19,DangLN19,Witt19,DoerrK20tec}.

The second function we consider in this paper is $\leadingones$. This function, first proposed by Rudolph~\cite{Rudolph97}, returns the length of the longest prefix consisting only of one-bits in the argument. Formally, it is defined by
\begin{align*}
	\leadingones(x) = \sum_{i = 1}^n \prod_{j = 1}^i x_j.
\end{align*}
This function is still considered to be easy for most standard EAs. Due to the low correlation between fitness and the distance to the optimum (e.g., we have all sub-optimal fitness levels in distance one from the optimum), the typical runtimes are higher than on \onemax, namely quadratic for many algorithms (when the parameters are set right)~\cite{Rudolph97,DrosteJW02,JansenJW05,Witt06,GutjahrS08,BottcherDN10,DoerrNSW11,Sudholt13,Doerr19tcs,DangLN19}. 
However, due to the low fitness-distance correlation, noise can have a drastic effect on the runtime, misleading the algorithms from good solutions~\cite{Sudholt21}. 

Finally, we test the algorithms on the $\jump_k$ function. This function from~\cite{DrosteJW02} has a positive integer parameter $k$ and is formally defined as follows.
\begin{align*}
	\jump_k = \begin{cases}
		\onemax(x) + k, &\text{ if } \onemax(x) \notin (n - k, n), \\
		n - \onemax(x), &\text{ otherwise.}
	\end{cases}
\end{align*}
This function generally imitates \onemax, but it has a valley of low fitness in a ball of radius $k-1$ around the optimum. Hence, it has a set of local optima in distance $k$ from the global one. This function is often used to analyze the ability of evolutionary algorithms to escape local optima~\cite{DrosteJW02,DoerrLMN17,CorusOY18fast,HasenohrlS18,RajabiW20,
Doerr21cgajump,BenbakiBD21,Doerr22,LissovoiOW23,DoerrDLS23}. Also, it is one of the few examples where crossover was shown to lead to super-constant speed-ups~\cite{JansenW02,FriedrichKKNNS16,DangFKKLOSS18,WhitleyVHM18,RoweA19,DoerrEJK23arxiv}.

\subsection{Noise Model}
We focus on the bitwise prior noise model. In this model we have a noise rate $q$ and the noise affects the individual \emph{before} we evaluate its fitness by flipping each bit independently with probability $q/n$.

The choice of $q$ in our experimental setup is mostly guided by the following theoretical results considering this noise model which have been mentioned in the introduction. In~\cite{GiessenK16} it was shown that the runtime of the \oea on \onemax is $O(n\log(n))$ (that is, same as without noise) if $q = O(\frac{1}{n})$, it is at most polynomial in $n$ if $q = O(\frac{\log(n)}{n})$ (for more precise bounds in this case see Corollary 14 in~\cite{Dang-NhuDDIN18}), and it is super-polynomial if $q = \omega(\frac{\log(n)}{n})$.

For this reason we consider $q = \frac{\ln(n)}{n}$, since it is the borderline value between the polynomial and super-polynomial runtimes of the \oea. We also consider higher noise rates such as $q = \frac{1}{6e} \approx 0.061$, which is a relatively small constant\footnote{The choice of this constant was guided by our preliminary theoretical analysis of the \ollga, from which we concluded that with noise rates up to this one we are very likely to have a beneficial mutation in the mutation winner.}, and $q = 1$, which is considered as a very high noise rate.

In~\cite{Sudholt21} it was shown that the runtime of the \oea on \leadingones with the bitwise noise is $\Theta(n^2) \cdot e^{\Theta(\min\{qn^2, n\})}$, which implies that any noise rate $q = \omega(\frac{\log(n)}{n^2})$ yields a super-polynomial runtime. Hence, we chose the same noise rates for our empirical investigation on \leadingones. We use the same noise rates for the experiments on \jump.

Some clues on how the non-trivial offspring population helps the optimization can be found in the results for the one-bit prior noise model, where we flip exactly one bit chosen uniformly at random with probability $q$ (which is also called the \emph{noise rate}) each time before evaluating fitness. For this noise model it was shown in~\cite{GiessenK16} that if we use the \oplea, then with $\lambda \ge \min\{12/q, 24\}n\ln(n)$ we can get a runtime of $O(n^2 \lambda/q)$ for any noise rate in $(0, 1)$. It was also shown in~\cite{Sudholt21} that the runtime of the \oplea on \leadingones with the one-bit noise is $O(n^2) \cdot e^{O(qn/\lambda)}$, which means that the larger population sizes can help to overcome large noise rates. We are, however, not aware of any theoretical results for the population-based EAs for the bitwise noise model.

\subsection{Algorithms}
\label{sec:algorithms}

In this paper we focus on the influence of the offspring population size on the runtime in a noisy environment. Hence, we consider two algorithms which create more than one offspring in each iteration. To minimize the effect of the parent population, all considered algorithms store only one individual and use it as a parent in each iteration. Due to the noisy environment, the fitness of this individual is recalculated in each iteration when we decide whether we should replace it with its offspring. This is a common practice and avoids that a single extreme noise event has a long-lasting impact on the optimization process~\cite{DoerrHK12ants}.

By the \emph{runtime} of an algorithm we understand the number of fitness evaluations made by the algorithm until it finds the optimal solution and accepts it as the current individual. We note that in practice it is hard to determine such a moment, since even if we find an individual with the best fitness, it might appear as sub-optimal individual due to noise. However, our main goal is to find the influence of the non-trivial offspring populations on the algorithm performance, hence in our experiments we use the knowledge of the true fitness of individuals to detect the moment of finding the truly best individual.

\subsubsection{The \oplea}
\label{sec:oplea}

We first consider a classic elitist mutation-based algorithm, the \oplea. This algorithm stores only one individual $x$, which is initialized with a random bit string. In each iteration we create $\lambda$ new individuals by flipping each bit of $x$ independently with probability $\frac{1}{n}$. We evaluate the fitness of all offspring and choose the one with the best value (the ties are broken uniformly at random). If the fitness of the chosen individual is not worse than the fitness of the current individual, we replace the current individual with the chosen one. The pseudo-code of the \oplea is shown in Algorithm~\ref{alg:oplea}. The typical runtime behavior of the \oplea is that for moderate population sizes, it has the same asymptotic runtime as the the \oea (this is called ``linear speed-up'' because the number of iterations reduces by a factor of~$\lambda$), but after a certain ``cut-off point'' the total work to solve a problem increases significantly~\cite{JansenJW05,NeumannW07,DoerrK13cec,DoerrK15}.

\begin{algorithm}[h]
    $x \gets $ random bit string of length $n$\;
    \While{not terminated}
        {
        \For{$i \in [1..\lambda]$}
            {$x^{(i)} \gets$ a copy of $x$\;
            Flip each bit in $x^{(i)}$ with probability $\frac{1}{n}$\;
            }
        $x' \gets \argmax_{z \in \{x^{(1)}, \dots, x^{(\lambda)}\}}f(z)$\;
        \If{$f(x') \ge f(x)$}
            {
             $x \gets x'$\;
            }
        }
        \caption{The \oplea maximizing a pseudo-Boolean function~$f: \{0,1\}^n \to \R$.}
    \label{alg:oplea}
\end{algorithm}

\subsubsection{The \ollga}
\label{sec:ollga}

The \ollga is a crossover-based algorithm which also stores only one individual~$x$ (initialized with a random bit string). This algorithm has three parameters: the population size $\lambda$, the mutation rate $p$ and the crossover bias $c$. Each iteration of the \ollga consists of two phases. The first phase is the \emph{mutation phase}, which starts with the choice of a number~$\ell$ from the binomial distribution $\Bin(n, p)$. Then we create $\lambda$ offspring, each by flipping exactly $\ell$ bits in $x$ (these bits are chosen uniformly at random). This can be interpreted as generating $\lambda$ offspring via the standard bit mutation with rate $p$, but conditional on that all of them have the same number of flipped bits. Then we choose the offspring with the best fitness as the mutation winner $x'$ (all ties are broken uniformly at random).

In the second phase, called the \emph{crossover phase}, we create another $\lambda$ offspring by applying a crossover operator to $x$ and $x'$. This crossover operator chooses each bit from $x'$ with probability $c$ and from $x$ with probability $(1 - c)$ (each bit is chosen independently from others). The best crossover offspring is chosen as the crossover winner $y$. If $y$ has a fitness which is not worse than the fitness of $x$, we replace $x$ with $y$. The pseudo-code of the \ollga is shown in Algorithm~\ref{alg:ollga}

\begin{algorithm}[h]
    $x \gets $ random bit string of length $n$\;
    \While{not terminated}
        {
        \nonl\textbf{Mutation phase:}\\
        Choose $\ell \sim \Bin\left(n, p\right)$\;
        \For{$i \in [1..\lambda]$}
            {$x^{(i)} \gets$ a copy of $x$\;
            Flip $\ell$ bits in $x^{(i)}$ chosen uniformly at random\;
            }
        $x' \gets \argmax_{z \in \{x^{(1)}, \dots, x^{(\lambda)}\}}f(z)$\;
        \nonl\textbf{Crossover phase:}\\
        \For{$i \in [1..\lambda]$}
            {Create $y^{(i)}$ by taking each bit from $x'$ with probability $c$ and from $x$ with probability $(1 - c)$\;
            }
        $y \gets \argmax_{z \in \{y^{(1)}, \dots, y^{(\lambda)}\} }f(z)$\;
        \If{$f(y) \ge f(x)$}
            {
             $x \gets y$\;
            }
        }
        \caption{The \ollga maximizing a pseudo-Boolean function~$f: \{0,1\}^n \to \R$.}
    \label{alg:ollga}
\end{algorithm}

The authors who first proposed the \ollga in~\cite{DoerrDE15} recommend to use $p = \frac{\lambda}{n}$ and $c = \frac{1}{\lambda}$. This setting assumes quite a strong mutation strength, therefore, the mutation offspring have a lot of bits flipped from the right position to the wrong one. However, this also maximizes our chances that in the best individual there is at least one beneficial bit flip. Then the biased crossover has a good chance to keep the beneficial bits and undo the destructive bit flips. We note that this setting works well for \onemax~\cite{DoerrDE15} and \leadingones~\cite{AntipovDK19foga}, but the most effective regime for $\jump_k$ is obtained when $p = c = \sqrt{\frac{k}{n}}$ and $\lambda = \frac{1}{\sqrt{n}} \sqrt{\frac{n}{k}}^k$, as it was shown in~\cite{AntipovDK22}.

\section{Results for \onemax}
\label{sec:onemax}
In this section we show the results of the \ollga and the \oplea optimizing a noisy \onemax function. We start with a discussion of what is the optimal population size for these algorithms in the presence of noise.

We considered two different problem sizes $n=2^7$ and $n=2^{10}$. We ran the \oplea and the \ollga with standard parameters $p=\frac{\lambda}{n}$ and $c=\frac{1}{\lambda}$ using all population sizes $\lambda \in [2..30]$ and tracked the mean runtime and its standard deviation over 128 runs for each setting. We used a setting without noise as a baseline and two different noise rates $q = \frac{\ln(n)}{n}$ and $q=\frac{1}{6e}$. The results of the experiments are provided in Figure~\ref{plot:optimal_lambda:n=128}.

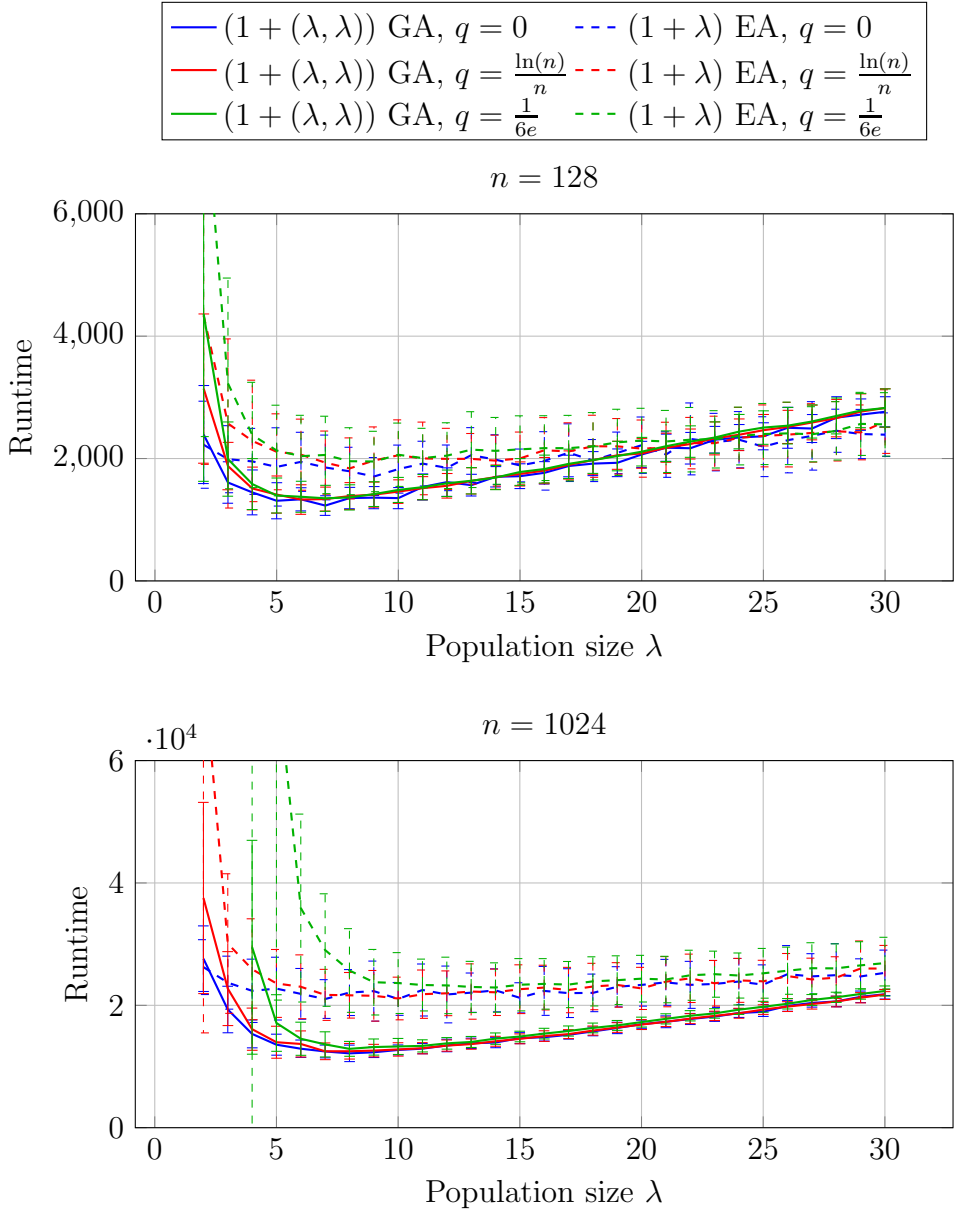
\begin{figure*}[!h]
\begin{tikzpicture}
    \begin{axis}[width=0.9\linewidth, height=0.31\textheight, name = axis1,
                    at={(0, 0)},anchor=outer south east,
                    legend style={at={(0.5,1.2)}, anchor=south,legend columns = 2},
                    ymin = 0, ymax = 6000,
                    legend cell align={left},
                    cycle list name=optlamruntimes, grid=major, log base x=2,
                    xlabel={Population size $\lambda$}, ylabel={Runtime}, title={$n = 128$}]

                    \addplot plot [error bars/.cd, y dir=both, y explicit] coordinates
                    {(2,2392.1875)+-(0,801.2343382829707)(3,1608.25)+-(0,338.8679683888697)(4,1446.75)+-(0,366.8742120945543)(5,1310.03125)+-(0,294.9605359593678)(6,1333.3125)+-(0,189.25320959959967)(7,1231.40625)+-(0,160.52275138103477)(8,1354.15625)+-(0,174.48873698877387)(9,1360.875)+-(0,181.0460490455398)(10,1353.1875)+-(0,172.76708263946)(11,1535.25)+-(0,213.2932253964012)(12,1614.84375)+-(0,233.00511117985695)(13,1565.15625)+-(0,177.37021828914092)(14,1701.9375)+-(0,166.50355429764855)(15,1711.78125)+-(0,202.50119357287133)(16,1766.53125)+-(0,182.80487964886905)(17,1877.96875)+-(0,176.2227788154457)(18,1917.0625)+-(0,196.3174943649954)(19,1930.5)+-(0,226.35964194175605)(20,2066.65625)+-(0,214.72695006900625)(21,2176.875)+-(0,222.07272879622118)(22,2168.4375)+-(0,258.12254084785)(23,2307.40625)+-(0,217.72701649298716)(24,2344.34375)+-(0,211.5065911170087)(25,2363.53125)+-(0,221.0606512327273)(26,2502.59375)+-(0,334.60759287699597)(27,2490.46875)+-(0,228.52912620372376)(28,2668.3125)+-(0,342.1298581587845)(29,2717.6875)+-(0,256.72522245340446)(30,2764.0625)+-(0,246.10566042606578)};
                    \addlegendentry{\ollga, $q = 0$};
                    \addplot plot [error bars/.cd, y dir=both, y explicit] coordinates
                    {(2,2225.7)+-(0,711.4225959301546)(3,1989.1)+-(0,548.0248078326382)(4,1953.375)+-(0,539.458510336986)(5,1863.0)+-(0,639.6236393380095)(6,1946.21875)+-(0,482.81762436600997)(7,1854.25)+-(0,529.2475200697685)(8,1791.5625)+-(0,392.4811251432991)(9,1705.9375)+-(0,310.4619478998191)(10,1827.375)+-(0,392.5065087039959)(11,1917.0)+-(0,365.68018814259)(12,1848.03125)+-(0,373.0868267219274)(13,2056.25)+-(0,448.652357065022)(14,1981.40625)+-(0,407.44676487970486)(15,1887.0)+-(0,268.3561066940717)(16,1961.375)+-(0,477.3783058277785)(17,2095.875)+-(0,471.27564585388876)(18,1967.09375)+-(0,343.54633597367547)(19,2092.5)+-(0,339.9908086992941)(20,2218.125)+-(0,460.67815161455184)(21,2065.25)+-(0,372.24680992588776)(22,2320.84375)+-(0,587.7489956060643)(23,2268.75)+-(0,469.91854347322794)(24,2307.8125)+-(0,459.51050297436075)(25,2193.75)+-(0,489.00351481354403)(26,2300.0625)+-(0,533.9091061161535)(27,2370.375)+-(0,559.5452254956698)(28,2454.125)+-(0,352.2597363806996)(29,2400.9375)+-(0,402.17657327814356)(30,2389.90625)+-(0,356.44603022188016)};
                    \addlegendentry{\oplea, $q = 0$};
            
                    \addplot plot [error bars/.cd, y dir=both, y explicit] coordinates
                    {(2,3143.310546875)+-(0,1219.7739797863223)(3,1877.0390625)+-(0,387.19574552042934)(4,1510.98046875)+-(0,349.30738893169763)(5,1411.201171875)+-(0,303.6002018720536)(6,1329.80859375)+-(0,241.4041656147372)(7,1337.34375)+-(0,195.77089956678316)(8,1386.6953125)+-(0,219.92911885827522)(9,1409.970703125)+-(0,204.80268019136886)(10,1459.6640625)+-(0,192.46363403834528)(11,1515.75390625)+-(0,188.52941887571873)(12,1554.39453125)+-(0,200.90791143293183)(13,1634.44921875)+-(0,214.04448226949148)(14,1696.443359375)+-(0,198.23479980654153)(15,1747.98828125)+-(0,196.9182091952669)(16,1803.076171875)+-(0,214.46109800432447)(17,1895.263671875)+-(0,229.9759642854386)(18,1974.513671875)+-(0,231.9698083815647)(19,2053.212890625)+-(0,229.33891829169943)(20,2083.47265625)+-(0,242.06296643863004)(21,2176.623046875)+-(0,237.1308160530102)(22,2237.431640625)+-(0,245.14822368721306)(23,2329.162109375)+-(0,269.3413390545082)(24,2388.462890625)+-(0,258.645145175575)(25,2462.94140625)+-(0,257.1035868658243)(26,2516.982421875)+-(0,270.99763909328533)(27,2589.189453125)+-(0,283.56801234119905)(28,2660.630859375)+-(0,305.7096867345292)(29,2767.23828125)+-(0,289.4055293347829)(30,2826.015625)+-(0,315.13575904815906)};
                    \addlegendentry{\ollga, $q = \frac{\ln(n)}{n}$};
                    \addplot plot [error bars/.cd, y dir=both, y explicit] coordinates
                    {(2,4335.6140625)+-(0,2431.10606825674)(3,2573.2)+-(0,1382.1711399099606)(4,2287.984375)+-(0,992.5994911120292)(5,2110.190625)+-(0,620.5817793910077)(6,2063.017578125)+-(0,581.0380440860861)(7,1933.3125)+-(0,515.9099023509337)(8,1838.56640625)+-(0,498.94104608180913)(9,1967.5390625)+-(0,549.3161572916102)(10,2053.712890625)+-(0,578.550608192647)(11,2001.9609375)+-(0,593.6931538885395)(12,1994.3828125)+-(0,499.4008030113086)(13,1994.7265625)+-(0,414.37962318016264)(14,1965.791015625)+-(0,415.78181872065545)(15,2001.375)+-(0,431.5600877919551)(16,2135.658203125)+-(0,533.668983871811)(17,2115.4921875)+-(0,468.7401414978931)(18,2200.400390625)+-(0,507.21169635737147)(19,2199.21875)+-(0,454.21740625876225)(20,2162.5078125)+-(0,467.2331498783181)(21,2174.77734375)+-(0,418.5379094827544)(22,2251.84375)+-(0,420.27089361914835)(23,2251.125)+-(0,440.85568429475876)(24,2338.76953125)+-(0,497.5730646816157)(25,2370.1640625)+-(0,506.04971774865766)(26,2390.607421875)+-(0,527.8708487800693)(27,2418.5)+-(0,477.10212481186875)(28,2445.572265625)+-(0,485.41040409397846)(29,2427.421875)+-(0,448.5371119221735)(30,2604.12109375)+-(0,520.1312343642744)};
                    \addlegendentry{\oplea, $q = \frac{\ln(n)}{n}$};
            
                    \addplot plot [error bars/.cd, y dir=both, y explicit] coordinates
                    {(2,4362.666015625)+-(0,2735.074512627821)(3,1991.158203125)+-(0,606.836060455084)(4,1582.62890625)+-(0,416.2876472562917)(5,1395.990234375)+-(0,285.9366494221361)(6,1375.2578125)+-(0,253.32104279000362)(7,1349.853515625)+-(0,212.36077246740481)(8,1365.046875)+-(0,205.78635801841767)(9,1418.6171875)+-(0,202.4961176747343)(10,1486.775390625)+-(0,204.82629440272962)(11,1532.5546875)+-(0,205.99171549610276)(12,1595.5078125)+-(0,216.48793796125187)(13,1636.6640625)+-(0,215.11147282222325)(14,1688.740234375)+-(0,197.45996652788907)(15,1780.078125)+-(0,218.9583957119123)(16,1828.40625)+-(0,227.49992058721142)(17,1913.10546875)+-(0,239.75273030529345)(18,1975.236328125)+-(0,217.95972608693637)(19,2040.416015625)+-(0,241.01136346990936)(20,2111.900390625)+-(0,233.7215721346607)(21,2201.23046875)+-(0,250.5093280631587)(22,2277.0703125)+-(0,257.26857572613164)(23,2343.0234375)+-(0,256.2261106194753)(24,2439.759765625)+-(0,280.8551167137962)(25,2506.072265625)+-(0,273.72770430813983)(26,2536.9609375)+-(0,298.6854703410782)(27,2605.732421875)+-(0,271.21687708837743)(28,2696.14453125)+-(0,293.63818119322934)(29,2784.75390625)+-(0,280.24611460351457)(30,2826.134765625)+-(0,311.7778584346575)};
                    \addlegendentry{\ollga, $q = \frac{1}{6e}$};
                    \addplot plot [error bars/.cd, y dir=both, y explicit] coordinates
                    {(2,8150.55)+-(0,5791.67396494744)(3,3228.10625)+-(0,1721.7098372579908)(4,2385.4921875)+-(0,860.4894319159096)(5,2131.040625)+-(0,740.5877553332957)(6,2038.96875)+-(0,667.5736433802173)(7,2058.65625)+-(0,632.9764563835985)(8,1956.19921875)+-(0,547.4311880267599)(9,1951.71875)+-(0,493.4589480376635)(10,2061.984375)+-(0,517.595933655887)(11,2007.2109375)+-(0,486.7217851405165)(12,2047.169921875)+-(0,537.6121834542363)(13,2145.58203125)+-(0,619.708050205598)(14,2125.458984375)+-(0,546.124962696468)(15,2150.3125)+-(0,548.0293352948818)(16,2170.189453125)+-(0,529.6694034944235)(17,2169.52734375)+-(0,535.9450648467334)(18,2200.95703125)+-(0,550.4039231532025)(19,2264.3359375)+-(0,541.084656866184)(20,2294.783203125)+-(0,526.51553005561)(21,2278.890625)+-(0,507.58569526446405)(22,2307.81640625)+-(0,526.634228686842)(23,2313.84375)+-(0,468.4275297054364)(24,2368.115234375)+-(0,471.89591468793606)(25,2400.68359375)+-(0,499.4038882879117)(26,2411.279296875)+-(0,510.3911433977676)(27,2396.5703125)+-(0,450.8682107818784)(28,2452.595703125)+-(0,476.8792662623443)(29,2566.40625)+-(0,515.4014430140232)(30,2558.287109375)+-(0,518.2109337567877)};
                    \addlegendentry{\oplea, $q = \frac{1}{6e}$};

    \end{axis}
    
    \begin{axis}[width=0.9\linewidth, height=0.31\textheight,
                      at={(0, -0.5)},anchor=outer north east,
                      ymin = 0, ymax = 60000,
                      legend cell align={left},
                      cycle list name=optlamruntimes, grid=major, log base x=2,
                      xlabel={Population size $\lambda$}, ylabel={Runtime}, title={$n = 1024$}]

                      \addplot plot [error bars/.cd, y dir=both, y explicit] coordinates
                      {(2,27622.03125)+-(0,5361.474240381412)(3,19315.625)+-(0,3801.231533578953)(4,15326.4375)+-(0,2274.6375389485133)(5,13567.125)+-(0,1715.917916706682)(6,12909.40625)+-(0,1420.188584734766)(7,12461.25)+-(0,1045.5710162394519)(8,12149.15625)+-(0,1347.294305018743)(9,12325.0625)+-(0,857.2721977841985)(10,12706.96875)+-(0,817.0349397507046)(11,12921.6875)+-(0,835.7192799282244)(12,13437.5)+-(0,1022.4625604392563)(13,13791.9375)+-(0,846.269052130438)(14,13922.71875)+-(0,848.818636192937)(15,14638.78125)+-(0,897.1184681514685)(16,14787.09375)+-(0,621.7873108716014)(17,15187.8125)+-(0,708.9086435103962)(18,15734.25)+-(0,712.7303574985424)(19,16300.78125)+-(0,926.0695753551336)(20,16916.34375)+-(0,663.8692740938817)(21,17308.84375)+-(0,964.5682359667135)(22,17766.5625)+-(0,902.4882595323609)(23,18158.15625)+-(0,778.4859949516995)(24,18739.4375)+-(0,720.3518210525673)(25,18944.90625)+-(0,756.0206412267706)(26,19899.84375)+-(0,873.7954962895709)(27,20384.375)+-(0,677.3532474824344)(28,20632.21875)+-(0,863.3294466184027)(29,21389.34375)+-(0,944.1134270234363)(30,21811.3125)+-(0,784.7362390279616)};
                      \addplot plot [error bars/.cd, y dir=both, y explicit] coordinates
                      {(2,26245.3125)+-(0,4482.831746211735)(3,23704.125)+-(0,4330.085159021125)(4,22370.9375)+-(0,5174.4331569838405)(5,22730.0625)+-(0,5130.449456050976)(6,21873.25)+-(0,4149.246678615288)(7,21027.0)+-(0,3168.8191807043836)(8,22087.125)+-(0,3754.583130571888)(9,22323.4375)+-(0,4907.1972712123315)(10,21038.1875)+-(0,2676.7318743840874)(11,22505.625)+-(0,4248.81240576411)(12,21774.59375)+-(0,4641.4705028375365)(13,22068.8125)+-(0,3211.7789797624228)(14,22448.4375)+-(0,3512.0565639798215)(15,21227.5)+-(0,2462.094179758362)(16,22356.0625)+-(0,3996.88478644228)(17,21993.75)+-(0,3986.9282583838904)(18,22034.65625)+-(0,3071.8666618338007)(19,23030.0)+-(0,3285.1712284141295)(20,23336.25)+-(0,3466.2088800301694)(21,23756.5625)+-(0,3754.788850067837)(22,23345.0)+-(0,3840.5222382509387)(23,23493.75)+-(0,3833.9406538834164)(24,23887.5)+-(0,3771.82192381878)(25,23377.25)+-(0,3234.8309751051906)(26,25093.96875)+-(0,4673.364142164982)(27,24731.0)+-(0,3709.9933962205378)(28,24933.65625)+-(0,5141.214877155587)(29,24720.9375)+-(0,2880.131680339243)(30,25309.5625)+-(0,3716.049741808329)};
              
                      \addplot plot [error bars/.cd, y dir=both, y explicit] coordinates
                      {(2,37543.193359375)+-(0,15625.12209351407)(3,22732.650390625)+-(0,6041.676710116999)(4,16099.681640625)+-(0,3447.6031113079757)(5,13968.23828125)+-(0,2613.8555099346627)(6,13686.359375)+-(0,2132.804984361758)(7,12487.470703125)+-(0,1365.7467047888833)(8,12493.140625)+-(0,1274.153232815174)(9,12600.080078125)+-(0,1021.8592905107429)(10,12792.1171875)+-(0,1091.871707716085)(11,12986.28515625)+-(0,915.887471304288)(12,13419.53125)+-(0,917.7805658174439)(13,13664.63671875)+-(0,836.7372371086597)(14,14124.076171875)+-(0,813.5604618325675)(15,14521.259765625)+-(0,805.537995266949)(16,14972.783203125)+-(0,867.0243933720897)(17,15295.478515625)+-(0,728.9085212500411)(18,15867.724609375)+-(0,748.1189329457441)(19,16371.697265625)+-(0,789.0910639337029)(20,16856.365234375)+-(0,809.2272605282006)(21,17304.64453125)+-(0,773.0096156125535)(22,17821.58203125)+-(0,814.4539828289097)(23,18274.921875)+-(0,765.710172398953)(24,18701.15625)+-(0,811.6766134395443)(25,19302.90234375)+-(0,783.022625387994)(26,19792.498046875)+-(0,800.6126706677114)(27,20214.6484375)+-(0,827.2214253511926)(28,20622.64453125)+-(0,873.9157403881783)(29,21239.76953125)+-(0,840.361720858438)(30,21768.421875)+-(0,812.4645699907377)};
                      \addplot plot [error bars/.cd, y dir=both, y explicit] coordinates
                      {(2,75737.90625)+-(0,60242.98597780874)(3,30095.4296875)+-(0,11407.659568508943)(4,25897.6953125)+-(0,8245.232823718308)(5,23576.3671875)+-(0,5549.756145221459)(6,23032.32421875)+-(0,5213.397736350278)(7,21707.71875)+-(0,4170.52744516787)(8,21638.3203125)+-(0,3724.7767601595806)(9,21556.54296875)+-(0,4138.010160451208)(10,21116.240234375)+-(0,3487.3017572047115)(11,21821.6953125)+-(0,3950.150167356366)(12,21967.791015625)+-(0,3657.4258693141896)(13,22252.91796875)+-(0,4522.7949298686835)(14,22130.419921875)+-(0,3793.6301038294764)(15,22635.65625)+-(0,3986.258475291829)(16,22900.759765625)+-(0,3653.3023479864173)(17,22496.73046875)+-(0,3706.7932366437576)(18,23072.75390625)+-(0,3793.3929905618947)(19,23333.90625)+-(0,3639.32669647985)(20,22764.328125)+-(0,3444.9194619562927)(21,23861.4921875)+-(0,4098.72894454811)(22,24414.3203125)+-(0,3955.673515332111)(23,23606.71875)+-(0,3726.574145028975)(24,24077.24609375)+-(0,3590.3812109884816)(25,23923.453125)+-(0,3438.4817549396325)(26,24778.037109375)+-(0,3692.349965822209)(27,24233.78125)+-(0,3472.9189743137454)(28,24525.390625)+-(0,3413.6229842591156)(29,26012.4609375)+-(0,4509.336126655632)(30,26013.84375)+-(0,3776.9174712933477)};
              
                      \addplot plot [error bars/.cd, y dir=both, y explicit] coordinates
                      {(4,29489.642578125)+-(0,17476.668083811135)(5,17109.46875)+-(0,4626.126798020909)(6,14537.173828125)+-(0,2707.0326306018474)(7,13617.626953125)+-(0,2054.4435177138976)(8,12886.697265625)+-(0,1216.928088592347)(9,13180.841796875)+-(0,1317.7850914460203)(10,13276.7578125)+-(0,1310.0596703138049)(11,13372.43359375)+-(0,993.8302952448471)(12,13783.837890625)+-(0,945.9289142583208)(13,14011.1015625)+-(0,899.2185960238248)(14,14556.52734375)+-(0,928.8356744816703)(15,14903.25)+-(0,865.6267779765134)(16,15373.875)+-(0,871.8278011418023)(17,15800.17578125)+-(0,826.932495378826)(18,16303.630859375)+-(0,796.9241298977739)(19,16716.0703125)+-(0,806.7958731882882)(20,17260.759765625)+-(0,807.4172430638449)(21,17788.814453125)+-(0,829.1903493735514)(22,18291.97265625)+-(0,838.8559541138868)(23,18677.083984375)+-(0,773.1185294149597)(24,19290.11328125)+-(0,779.9766641011502)(25,19757.51953125)+-(0,894.1239119557368)(26,20255.5234375)+-(0,852.8239163292641)(27,20873.681640625)+-(0,888.3785100749192)(28,21310.986328125)+-(0,833.4470388752694)(29,21850.05078125)+-(0,844.6663966561086)(30,22343.751953125)+-(0,823.5513553853959)};
                      \addplot plot [error bars/.cd, y dir=both, y explicit] coordinates
                      {(4,701132.421875)+-(0,771353.3021264387)(5,74965.37109375)+-(0,54108.029607614975)(6,35897.39453125)+-(0,15356.952232767455)(7,29033.640625)+-(0,9195.201693456735)(8,25691.765625)+-(0,6840.021809272128)(9,23777.40234375)+-(0,5356.8858004202875)(10,23638.35546875)+-(0,4974.642784620768)(11,23319.1640625)+-(0,4234.922979814213)(12,23252.353515625)+-(0,4616.459250735197)(13,23002.08203125)+-(0,4183.701075433494)(14,22909.541015625)+-(0,4034.476656140511)(15,23363.09375)+-(0,4273.623395751074)(16,23504.16015625)+-(0,4127.573814657192)(17,23390.12109375)+-(0,3774.337173931842)(18,23915.28515625)+-(0,4016.5515643978933)(19,24108.515625)+-(0,3983.3492483249393)(20,24356.47265625)+-(0,3813.426406302286)(21,24213.921875)+-(0,3797.8501826061392)(22,24914.66015625)+-(0,4090.142597812908)(23,25091.71875)+-(0,3764.575065116704)(24,24869.970703125)+-(0,3684.290227650852)(25,25233.0)+-(0,4064.808144303984)(26,25659.544921875)+-(0,3854.20068494844)(27,26067.3984375)+-(0,4161.751814965972)(28,26036.845703125)+-(0,4008.0811330074894)(29,26532.1875)+-(0,3863.3813087396315)(30,26900.916015625)+-(0,4219.281668671487)};

    \end{axis}

\end{tikzpicture}
\caption{Mean runtimes (number of fitness evaluation) and their standard deviation of $128$ runs of the \ollga with standard parameters $p=\frac{\lambda}{n}$, $c=\frac{1}{\lambda}$ and the \oplea with different noise rates on \onemax with varying population size $\lambda$ for the problem sizes $n=2^7$ and $n=2^{10}$.}
\label{plot:optimal_lambda:n=128}
\end{figure*}

The data in the plots suggests that with too small values of $\lambda$ the runtime of both algorithms is extremely large, especially for the large noise rates. The optimal choice of $\lambda$ for the \ollga for all noise rates seems to be $\lambda = 7$ for $n = 128$ and $\lambda = 8$ for $n = 1024$. For the \oplea it is not so clear, which values of $\lambda$ are better due to the larger standard deviations of the runtimes, but it seems like $\lambda = 8$ for $n = 128$ and $\lambda = 10$ for $n = 1024$ is the most balanced choice for all noise rates. The observation that this value is slightly larger for both algorithms for $n=1024$ compared with $n = 128$ makes us assume that the optimal value of $\lambda$ grows with the growth of $n$, but very slowly. We also note that choosing $\lambda$ slightly smaller than these optimal values can drastically increase the runtime, while the choice of a too large $\lambda$ is not so critical.


In this section we also compare the runtimes of the \oplea and the \ollga with standard parameters $p=\frac{\lambda}{n}$ and $c=\frac{1}{\lambda}$ for varying problem size $n$. We made $100$ runs of each algorithm for problems sizes $\{2^5, \dots, 2^{14}\}$, for which the runs took a reasonable time. We show the results of our experiments in Figure~\ref{plot:onemax_normalized}, where we normalize the runtimes by $n\ln(n)$, which is asymptotically the same as the runtime of both algorithms with logarithmic $\lambda$. This normalization allows us to better show how the ratio of the runtimes changes with the growth of the problem size. We use the same noise rates $q = \frac{\ln(n)}{n}$ and $q=\frac{1}{6e}$ as before, and also use the setting without noise ($q = 0$) as a baseline and a setting with a very strong noise, $q = 1$. For the population size we took $\lambda = \ln(n)$, which is close to the optimal value observed in the previous experiment for both $n = 128$ and $n=1024$. We also took a slightly smaller population size, $\lambda = \ln(n) / 2$, and significantly larger one, $\lambda = \sqrt{n}$.


The results of the experiments show that both algorithms withstand all noise rates up to $q = \frac{1}{6e}$, when the population size is at least $\ln(n)$. For $q=1$, however, it is necessary to use $\lambda = \sqrt{n}$ to obtain a reasonable runtime. A smaller population size $\lambda = \frac{\ln(n)}{2}$ yields a poor performance of both algorithms when the noise rate is $\frac{1}{6e}$, but it makes both algorithms sustainable to the smaller noise rate $q = \frac{\ln(n)}{n}$. When the population size is equal for both algorithms and less than $\sqrt{n}$, then the \ollga always has an advantage over the \oplea. This means that its more complex mechanics do not render it unstable under noise, while maintaining its better performance which was observed in the setting with no noise. On large population size $\lambda=\sqrt{n}$ the \oplea becomes better than the \ollga on sufficiently large problem sizes for all noise levels, except $q=1$. 

We also note that at $q= \frac{1}{6e}$ the \ollga is more effective with $\lambda = \ln(n)$ than with larger $\lambda = \sqrt{n}$, while for the \oplea it is already better to choose $\lambda =\sqrt{n}$ than $\lambda =\ln(n)$. This observation indicates for this particular noise rate that the core mechanisms of the \ollga which rely on the intermediate selection are more robust to noise than the simple mechanisms of the \oplea, which needs a larger population size to reduce the effect of the noise.

\begin{figure*}[!h]
\begin{tikzpicture}
    \begin{axis}[width=0.5\linewidth, height=0.26\textheight, name = axis1,
                        at={(0, 0)},anchor=outer south east,
                        cycle list name=onemaxruntimes, grid=major,  xmode=log, log base x=2,
                        xlabel={Problem size $n$}, ylabel={Runtime $/(n\ln(n))$}, title={$q = 0$}]

                        \addplot plot [error bars/.cd, y dir=both, y explicit] coordinates
                        {(32,6.45912603494)+-(0,2.560025478293311)(64,3.840611993616518)+-(0,1.3331655952966228)(128,3.8078454265606143)+-(0,1.1453797691247982)(256,3.859948897362808)+-(0,1.0146761859782911)(512,2.568416706496502)+-(0,0.4245874860492325)(1024,2.5888105847345546)+-(0,0.44311940130431204)(2048,2.6436720607403474)+-(0,0.4217927546261373)(4096,2.1158271013365484)+-(0,0.2865251424746304)(8192,2.1522278675795907)+-(0,0.3003517900591458)(16384,2.0744591846354443)+-(0,0.26128877427170955)};
                        \addplot plot [error bars/.cd, y dir=both, y explicit] coordinates
                        {(32,4.5881309037871265)+-(0,1.613974132951132)(64,3.547564563435947)+-(0,1.1660900546710902)(128,3.6389245150208143)+-(0,1.2429250026781684)(256,3.643727795873326)+-(0,1.0504938941923074)(512,3.250234132352737)+-(0,0.7014507857197395)(1024,3.3603861529356114)+-(0,0.6577238225793387)(2048,3.3270463722250674)+-(0,0.5084941135713912)(4096,3.278068941946972)+-(0,0.5798770271768644)(8192,3.2009322076784885)+-(0,0.4088696613118337)(16384,3.1856951405145884)+-(0,0.43447832913530887)};
                        \addplot plot [error bars/.cd, y dir=both, y explicit] coordinates
                        {(32,2.7102829711900283)+-(0,1.0371645581000162)(64,2.3527200221497053)+-(0,0.6182389327934957)(128,2.3300652015857453)+-(0,0.388564558105513)(256,2.0298113406101095)+-(0,0.30594002736633014)(512,1.8516527484215117)+-(0,0.20013804519771913)(1024,1.8146595804644112)+-(0,0.19072574364624428)(2048,1.683482559487617)+-(0,0.14762366282377398)(4096,1.6205839887407631)+-(0,0.17144688910534625)(8192,1.5315975192974094)+-(0,0.11524402573395787)(16384,1.4651041876758186)+-(0,0.09648546633729801)};
                        \addplot plot [error bars/.cd, y dir=both, y explicit] coordinates
                        {(32,3.333707565734173)+-(0,1.4721345222537208)(64,3.2026702802234293)+-(0,1.1124628833063237)(128,3.0353756209060587)+-(0,0.9251541643670588)(256,3.1425955570364126)+-(0,0.7946829991487842)(512,3.0018701735774784)+-(0,0.7224856941618839)(1024,3.1134288844421905)+-(0,0.6078552920684672)(2048,3.0172819181864696)+-(0,0.49604982406284565)(4096,3.023095076876451)+-(0,0.4683722016047373)(8192,2.9337458880847676)+-(0,0.3541608506940306)(16384,2.9429095712556133)+-(0,0.3234622552605031)};
                        \addplot plot [error bars/.cd, y dir=both, y explicit] coordinates
                        {(32,2.5064121282244076)+-(0,0.7857396603534904)(64,2.4417237865212145)+-(0,0.48328899938944225)(128,2.4034719726952636)+-(0,0.31048536701958757)(256,2.5471245877913686)+-(0,0.22691908493423207)(512,2.7845141394657684)+-(0,0.15323195145842253)(1024,3.1871655351795)+-(0,0.12577440898957323)(2048,3.7123281756119035)+-(0,0.09571538450141313)(4096,4.498018290675898)+-(0,0.08312505242178382)(8192,5.4597621741097635)+-(0,0.05357173771398159)(16384,6.791748994751615)+-(0,0.05338270999832455)};
                        \addplot plot [error bars/.cd, y dir=both, y explicit] coordinates
                        {(32,2.882504691696149)+-(0,1.1333903052048504)(64,2.9346445921582767)+-(0,0.8185202101834461)(128,3.0298528039526564)+-(0,0.7554886208439268)(256,3.1426096458551713)+-(0,0.6005927440013429)(512,3.331376335856902)+-(0,0.5747917328630738)(1024,3.7626402439603486)+-(0,0.6084205678448049)(2048,4.004167809784343)+-(0,0.48272120128154794)(4096,4.391609551830382)+-(0,0.40240005040448057)(8192,4.968008076197918)+-(0,0.4108786778894578)(16384,5.651838833390377)+-(0,0.3949194926943413)};

    \end{axis}

    \begin{loglogaxis}[width=0.5\linewidth, height=0.26\textheight, name=axis2,
                        at={(0.5\linewidth, 0)},anchor=outer south east,
                        legend style={at={(-0.1,1.3)}, anchor=south,legend columns = 2},
                        legend cell align={left},
                        cycle list name=onemaxruntimes, grid=major, log base x=2,
                        xlabel={Problem size $n$}, 
                        title={$q = \frac{\ln n}{n}$}]

                        \addplot plot [error bars/.cd, y dir=both, y explicit] coordinates
                        {(32,20.6430724978799)+-(0,16.275403324873732)(64,5.2292059704721465)+-(0,2.739343882547369)(128,5.536422726947163)+-(0,2.6156393352861436)(256,5.421694456762627)+-(0,2.3265282225363495)(512,2.931717248705084)+-(0,0.8524098576684075)(1024,3.0066412737788903)+-(0,0.7458171548323133)(2048,2.876621072902353)+-(0,0.6129693595622654)(4096,2.2550211084676293)+-(0,0.44046353108920094)(8192,2.2240816560653323)+-(0,0.33818628888016794)(16384,2.2197701291752616)+-(0,0.43755725193512884)};
                        \addlegendentry{\ollga, $\lambda=\frac{\ln n}{2}$};
                        \addplot plot [error bars/.cd, y dir=both, y explicit] coordinates
                        {(32,14.652912519668869)+-(0,10.997274593397792)(64,5.672496563895293)+-(0,3.164040738539297)(128,7.015507174001405)+-(0,4.052264455356617)(256,7.439171036549511)+-(0,4.231347001665551)(512,4.391015179788999)+-(0,1.6254526521459252)(1024,4.3669533558583415)+-(0,1.65473774051173)(2048,4.543692720139517)+-(0,1.7071244490870034)(4096,3.6171383791539484)+-(0,0.7453341256998331)(8192,3.599011104357973)+-(0,0.7408273352816613)(16384,3.694975748315005)+-(0,0.9258825538392977)};
                        \addlegendentry{\oplea, $\lambda=\frac{\ln n}{2}$};
                        \addplot plot [error bars/.cd, y dir=both, y explicit] coordinates
                        {(32,3.2095456237776654)+-(0,1.375228270165349)(64,2.417979430639917)+-(0,0.7545185595223197)(128,2.3693367775313727)+-(0,0.4955005669389123)(256,2.0801013791692218)+-(0,0.34612010994424647)(512,1.8792073470648787)+-(0,0.26355186030488187)(1024,1.8571148269118216)+-(0,0.2154395252520951)(2048,1.699044300207433)+-(0,0.17996294437404955)(4096,1.619845499824162)+-(0,0.15895492494035862)(8192,1.5412059582214193)+-(0,0.11689097746986704)(16384,1.4785901433639552)+-(0,0.1079351005891698)};
                        \addlegendentry{\ollga, $\lambda=\ln n$};
                        \addplot plot [error bars/.cd, y dir=both, y explicit] coordinates
                        {(32,4.222047037161551)+-(0,2.1924801054460605)(64,3.6409264690768093)+-(0,1.3894705918924348)(128,3.4136000239248285)+-(0,1.0970589680663008)(256,3.2387517450644117)+-(0,0.8303576080805)(512,3.1425047624266345)+-(0,0.8207955251460288)(1024,3.217504397494444)+-(0,0.6325440310164321)(2048,3.0391323952794784)+-(0,0.4761806855939063)(4096,3.0902194927498425)+-(0,0.5094829319209291)(8192,2.961714902708972)+-(0,0.3979012473988167)(16384,3.0004837806990126)+-(0,0.34669724749653763)};
                        \addlegendentry{\oplea, $\lambda=\ln n$};
                        \addplot plot [error bars/.cd, y dir=both, y explicit] coordinates
                        {(32,2.7831390707549217)+-(0,0.8536644096768441)(64,2.403402199497601)+-(0,0.4594734490664632)(128,2.462725519017489)+-(0,0.350294310856943)(256,2.5357337778249747)+-(0,0.2016611162746619)(512,2.7608449239511845)+-(0,0.14719842494460209)(1024,3.2041073397368147)+-(0,0.12366180551182446)(2048,3.733016325157321)+-(0,0.09009043963175611)(4096,4.49143000680387)+-(0,0.08867270404479445)(8192,5.459958333817095)+-(0,0.054788397406824486)(16384,6.800283799886784)+-(0,0.05046524494331019)};
                        \addlegendentry{\ollga, $\lambda=\sqrt{n}$};
                        \addplot plot [error bars/.cd, y dir=both, y explicit] coordinates
                        {(32,3.5609320346741846)+-(0,1.745702767083945)(64,3.282920191872878)+-(0,1.0605168645312681)(128,3.1338041341309957)+-(0,0.852508740299705)(256,3.2113489925787775)+-(0,0.7184234064182521)(512,3.376382284669356)+-(0,0.6271109452937763)(1024,3.770358098876354)+-(0,0.6381531370830418)(2048,4.004167809784343)+-(0,0.4989561560914487)(4096,4.406529023861711)+-(0,0.39319860604188656)(8192,4.968008076197918)+-(0,0.4108786778894578)(16384,5.641323542875416)+-(0,0.38256969700256205)};
                        \addlegendentry{\oplea, $\lambda=\sqrt{n}$};

    \end{loglogaxis}

    \begin{loglogaxis}[width=0.5\linewidth, height=0.26\textheight, name=axis3,
                        at={(0, -0.29\textheight)},anchor=outer south east,
                        cycle list name=onemaxruntimes, grid=major, log base x=2,
                        xlabel={Problem size $n$}, ylabel={Runtime $/(n\ln(n))$}, title={$q = \frac{1}{6e}$}]

                        \addplot plot [error bars/.cd, y dir=both, y explicit] coordinates
                        {(32,12.953958772142002)+-(0,7.693536802429299)(64,5.137346872165543)+-(0,2.5642200338534673)(128,7.221928495700028)+-(0,4.368024679524036)(256,16.276424869388624)+-(0,14.358899884809457)(512,6.933589861833467)+-(0,4.636690346997678)(1024,42.008370249612945)+-(0,43.05051280438467)(2048,9385.971179990434)+-(0,9353.876649136562)(4096,1627.8585521827947)+-(0,1740.8947147012536)};
                        \addplot plot [error bars/.cd, y dir=both, y explicit] coordinates
                        {(32,8.550312496708552)+-(0,5.435089489860852)(64,5.550205617069939)+-(0,3.287389510613775)(128,11.828376474024154)+-(0,8.97268317809782)(256,75.83018580516585)+-(0,82.40793291256948)(512,76.1098152009031)+-(0,71.19632602483652)(1024,25907.257871976555)+-(0,25298.119888622987)};
                        \addplot plot [error bars/.cd, y dir=both, y explicit] coordinates
                        {(32,3.136328850452551)+-(0,1.3432907053056327)(64,2.424065800343667)+-(0,0.7144857788385115)(128,2.4817897034863785)+-(0,0.5770931212045963)(256,2.22446246058005)+-(0,0.41749588082336364)(512,1.9868208755922998)+-(0,0.2834354962547327)(1024,2.0719552241629526)+-(0,0.3622875357512793)(2048,1.868801758232489)+-(0,0.2636793075374317)(4096,1.808464550745985)+-(0,0.2364264122993785)(8192,1.6924236884077457)+-(0,0.16251047507431135)(16384,1.8132055640513058)+-(0,0.39604515085766995)};
                        \addplot plot [error bars/.cd, y dir=both, y explicit] coordinates
                        {(32,3.7499250850306383)+-(0,1.954282438513512)(64,3.6185722099797006)+-(0,1.4132174539728062)(128,3.6933637107043573)+-(0,1.2539307117876404)(256,4.124318536960083)+-(0,1.4888294632511774)(512,3.7659036840121507)+-(0,1.1653639726126106)(1024,5.027104683142613)+-(0,2.072907338982121)(2048,5.4575831609469585)+-(0,2.3205412166703403)(4096,5.987476762678441)+-(0,3.1842975652928165)(8192,6.638452259056244)+-(0,3.506242715105725)(16384,40.35393246470439)+-(0,35.74364998651336)};
                        \addplot plot [error bars/.cd, y dir=both, y explicit] coordinates
                        {(32,2.616507793532246)+-(0,0.76311845216002)(64,2.407873051317023)+-(0,0.483765241265267)(128,2.4186556939403334)+-(0,0.31488008379522264)(256,2.609192878832739)+-(0,0.22336342966881143)(512,2.8336841169335663)+-(0,0.18081674742769013)(1024,3.2856111562557864)+-(0,0.12426593801874183)(2048,3.8200231062032635)+-(0,0.08174301729451419)(4096,4.624975278163892)+-(0,0.08361875698680916)(8192,5.623481969842076)+-(0,0.0625737311980131)(16384,6.988324307722827)+-(0,0.05213719590123116)};
                        \addplot plot [error bars/.cd, y dir=both, y explicit] coordinates
                        {(32,2.9717714473511534)+-(0,1.1089861713510745)(64,3.224085284736625)+-(0,1.047655779323804)(128,3.223070889786)+-(0,0.910386108365915)(256,3.390502411914168)+-(0,0.8545979054669048)(512,3.606956761625321)+-(0,0.6019000762591813)(1024,3.8494428652144586)+-(0,0.6031321725699708)(2048,4.237920525016446)+-(0,0.5129630249580671)(4096,4.595178307040453)+-(0,0.4005207893392115)(8192,5.192853295870058)+-(0,0.34857708541056964)(16384,6.018559590099657)+-(0,0.4123598586178431)};

    \end{loglogaxis}

    \begin{loglogaxis}[width=0.5\linewidth, height=0.26\textheight, name=axis4,
        at={(0.5\linewidth, -0.29\textheight)},anchor=outer south east,
        cycle list name=onemaxruntimes, grid=major, log base x=2,
        xlabel={Problem size $n$}, title={$q = 1$}]

        \addplot plot [error bars/.cd, y dir=both, y explicit] coordinates
		{(32,840.1981806079556)+-(0,146.3721406934118)(64,3454.287378558555)+-(0,1820.0323380685838)(128,16883.653964231984)+-(0,0.0)};
		\addplot plot [error bars/.cd, y dir=both, y explicit] coordinates
		{(32,567.2559681803332)+-(0,93.0484956256343)(64,2941.1735761921504)+-(0,107.01825281748017)(128,10130.19237853919)+-(0,0.0)};
		\addplot plot [error bars/.cd, y dir=both, y explicit] coordinates
		{(32,12.942958222455223)+-(0,9.357560649847585)(64,17.146655982065468)+-(0,15.39194399243783)(128,173.1986633511077)+-(0,154.94971727623806)(256,628.5000281022739)+-(0,608.3808296973491)(512,7983.576775380551)+-(0,7687.468248470215)};
		\addplot plot [error bars/.cd, y dir=both, y explicit] coordinates
		{(32,80.2354846990397)+-(0,62.847206184854144)(64,708.0476755362866)+-(0,682.9341631110789)(128,16502.725715624583)+-(0,2075.392092817919)};
		\addplot plot [error bars/.cd, y dir=both, y explicit] coordinates
		{(32,6.127666849295761)+-(0,3.440101990521246)(64,4.721219521309133)+-(0,1.7474223484875326)(128,4.443645639502382)+-(0,1.3409950135041468)(256,4.181589585214123)+-(0,0.6021465204828278)(512,4.426284189414899)+-(0,0.39271204341268456)(1024,4.96632974566329)+-(0,0.28454090776734686)};
		\addplot plot [error bars/.cd, y dir=both, y explicit] coordinates
		{(32,20.550830183703066)+-(0,15.228057094451202)(64,19.204187073583284)+-(0,14.008197546509685)(128,21.85489768860946)+-(0,13.635000568724942)(256,16.911316353520462)+-(0,11.689168377969382)(512,11.998729972436463)+-(0,5.551959142642587)(1024,8.898407759543042)+-(0,2.2411000112500123)};

    \end{loglogaxis}
\end{tikzpicture}
\caption{Mean runtimes (number of fitness evaluation) and their standard deviation over $100$ runs of the \ollga  with standard parameters $p=\frac{\lambda}{n}$, $c=\frac{1}{\lambda}$ and the \oplea on \onemax with noise rates $q \in \{0, \frac{\ln(n)}{n}, \frac{1}{6e}, 1\}$ with varying problem size $n$ normalized by $n \ln n$.}
\label{plot:onemax_normalized}
\end{figure*}
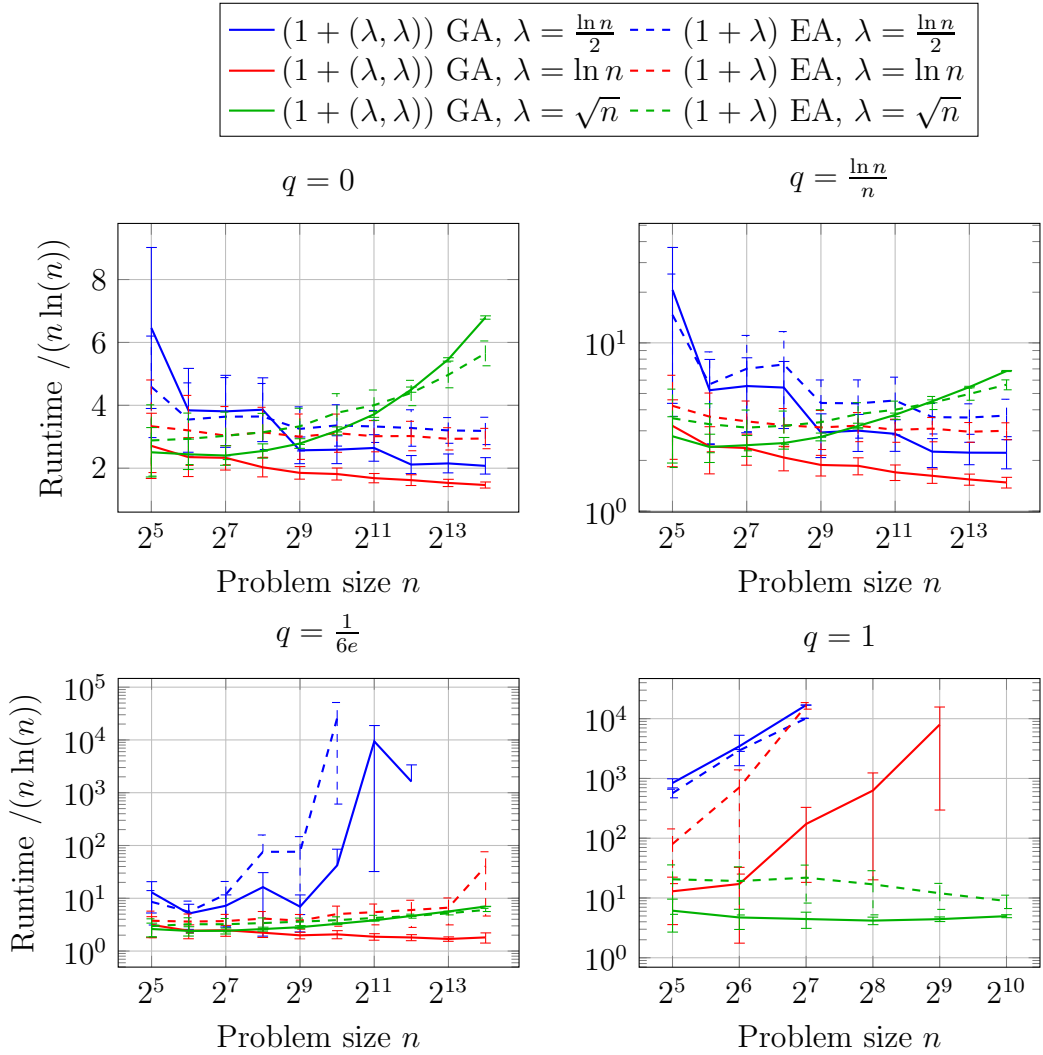

\section{Results for \leadingones}
\label{sec:leadingones}
In this section we discuss our results for the \leadingones benchmark.

As in Section \ref{sec:onemax}, we recorded the runtime of the algorithms for different noise levels and different population sizes. We took the same noise rates $q=0$, $q=\frac{\ln n}{n}$, $q=\frac{1}{6e}$ and $q = 1$. We also considered the same population sizes as for \onemax, that are, $\lambda = \frac{\ln(n)}{2}$, $\lambda = \ln(n)$ and $\lambda = \sqrt{n}$, but we also added the value $\lambda=\frac{n}{2}$. This additional value is motivated by the results in~\cite{Sudholt21}, which show that the expected runtime of the \oplea on \leadingones with prior one-bit noise with rate $q$ is $O(n^2) \cdot e^{O(qn/\lambda)}$. Hence, with $q = \Omega(1)$ we need to use $\lambda = \Omega(n)$ to make the exponential factor a constant. Since the one-bit noise model with rate $q$ is very similar to the bitwise noise model with the same rate, we assumed that we also need a linear value of $\lambda$ to be robust to the constant noise rates.
For the \ollga we used the standard parameters, which are $p=\frac{\lambda}{n}$ and $c=\frac{1}{\lambda}$. We made $100$ runs for each setting on problem sizes from $\{2^3, \dots, 2^9\}$, on which they did not take too much time. The results of the experiment are shown in Figure~\ref{plot:leadingones_normalized}, where the runtimes are normalized by $n^2$ (which is the suggested asymptotic runtime from~\cite{Sudholt21}) so that it was easier to see how the ratio between them changes with the problem size.

We observe that for increasing noise intensities, increasing population sizes are necessary to obtain a runtime which is not much larger than $n^2$ for the considered $n$. Once the population size is too small, the runtime drastically increases. 

Comparing the \oplea with the \ollga we see that in most settings the \oplea has a better performance than the \ollga. However, the advantage is usually at most a factor of two. This fits to the observation made in~\cite{AntipovDK19foga} that the \ollga does not gain from its working principles on a problem like \leadingones. So the higher cost of one iteration (twice as much as for the \oplea with same population size) does not amortize, but leads to twice as large runtimes. We note that for settings where the algorithms suffer strongly from the noise (that are, the logarithmic values of $\lambda$ with $q = \frac{\ln(n)}{n}$ and $q = \frac{1}{6e}$ and all sub-linear values of $\lambda$ with $q = 1$), the advantage of the \oplea vanishes.


\begin{figure*}[!h]
\begin{tikzpicture}
    \begin{axis}[width=0.5\linewidth, height=0.26\textheight, name = axis1,
                      at={(0, 0)},anchor=outer south east,
                      cycle list name=loruntimes, grid=major, xmode=log, log base x=2,
                      xlabel={Problem size $n$}, ylabel={Runtime $/n^2$}, title={$q = 0$}]

                      \addplot plot [error bars/.cd, y dir=both, y explicit] coordinates
                      {(8,2.6165625)+-(0,1.5807147994953896)(16,2.4533203125)+-(0,1.2028374979895764)(32,2.44365234375)+-(0,0.7070891062632538)(64,1.85670166015625)+-(0,0.39298650254028616)(128,1.8461761474609375)+-(0,0.3024865224116347)(256,1.8127403259277344)+-(0,0.1937608850090119)(512,1.6908411026000976)+-(0,0.14991183669223354)};
                      \addplot plot [error bars/.cd, y dir=both, y explicit] coordinates
                      {(8,1.585)+-(0,1.0379995032513263)(16,1.64890625)+-(0,0.7639297398720235)(32,1.69296875)+-(0,0.45439930595022576)(64,1.30679443359375)+-(0,0.2627325042263265)(128,1.2760125732421874)+-(0,0.21523361581419237)(256,1.283716278076172)+-(0,0.15009121588859628)(512,1.1401498413085938)+-(0,0.08793576233906508)};
                      \addplot plot [error bars/.cd, y dir=both, y explicit] coordinates
                      {(8,2.01171875)+-(0,1.1370006093851435)(16,1.924609375)+-(0,0.9620976387327214)(32,1.694013671875)+-(0,0.5286904472379862)(64,1.70246337890625)+-(0,0.4304765472744415)(128,1.691158447265625)+-(0,0.24503720180182653)(256,1.6827670288085939)+-(0,0.2051136185537253)(512,1.6996965408325195)+-(0,0.14676544794298452)};
                      \addplot plot [error bars/.cd, y dir=both, y explicit] coordinates
                      {(8,1.24265625)+-(0,0.9581207573903132)(16,1.291171875)+-(0,0.5452440579810769)(32,1.14203125)+-(0,0.3182146047370685)(64,1.0754150390625)+-(0,0.23725074060684237)(128,1.075225830078125)+-(0,0.1776938785277927)(256,1.0494351196289062)+-(0,0.11008687471690697)(512,1.0126656341552733)+-(0,0.08790986783187422)};
                      \addplot plot [error bars/.cd, y dir=both, y explicit] coordinates
                      {(8,2.01171875)+-(0,1.1370006093851435)(16,1.6846875)+-(0,0.7501668271489266)(32,1.648818359375)+-(0,0.5526362078621015)(64,1.762255859375)+-(0,0.43920232915156476)(128,1.841375732421875)+-(0,0.3226165034056742)(256,1.8970367431640625)+-(0,0.21103406891063872)(512,1.9633186340332032)+-(0,0.1423976301887155)};
                      \addplot plot [error bars/.cd, y dir=both, y explicit] coordinates
                      {(8,1.24265625)+-(0,0.9581207573903132)(16,1.0775390625)+-(0,0.460606798410205)(32,1.04185546875)+-(0,0.3083609256942181)(64,0.99322998046875)+-(0,0.20034760384786815)(128,0.96533203125)+-(0,0.1526892163574088)(256,0.9350933837890625)+-(0,0.10206624130888438)(512,0.9122641372680664)+-(0,0.06584377218787965)};
                      \addplot plot [error bars/.cd, y dir=both, y explicit] coordinates
                      {(8,1.65234375)+-(0,0.918567623799243)(16,1.868671875)+-(0,0.7963225298415231)(32,1.979677734375)+-(0,0.6400726720409347)(64,2.09361572265625)+-(0,0.44238914822047753)(128,2.1571124267578123)+-(0,0.30059679613521634)(256,2.202123260498047)+-(0,0.2196460667493484)(512,2.2916742324829102)+-(0,0.17689011092094106)};
                      \addplot plot [error bars/.cd, y dir=both, y explicit] coordinates
                      {(8,1.15078125)+-(0,0.6700424681911495)(16,1.0996875)+-(0,0.4368743713112429)(32,0.97982421875)+-(0,0.3025857819549428)(64,0.99652587890625)+-(0,0.22390705961678672)(128,1.015863037109375)+-(0,0.15505013261287012)(256,1.0024388122558594)+-(0,0.09935868508912461)(512,0.9925436782836914)+-(0,0.07806155718097509)};

    \end{axis}

    \begin{loglogaxis}[width=0.5\linewidth, height=0.26\textheight, name=axis2,
                        at={(0.5\linewidth, 0)},anchor=outer south east,
                        legend style={at={(-0.1,1.3)}, anchor=south,legend columns = 2},
                        legend cell align={left},
                        cycle list name=loruntimes, grid=major, log base x=2,
                        xlabel={Problem size $n$}, 
                        title={$q = \frac{\ln n}{n}$}]

                        \addplot plot [error bars/.cd, y dir=both, y explicit] coordinates
                        {(8,5.07890625)+-(0,4.625545457172287)(16,10.47984375)+-(0,10.51108315908718)(32,137.2875)+-(0,146.29281010537883)(64,73.145166015625)+-(0,65.52617263758924)};
                        \addlegendentry{\ollga, $\lambda=\frac{\ln n}{2}$};
                        \addplot plot [error bars/.cd, y dir=both, y explicit] coordinates
                        {(8,3.2665625)+-(0,2.653058579115574)(16,6.78484375)+-(0,6.0024816258731954)(32,111.71845703125)+-(0,115.53774217829694)(64,108.48951416015625)+-(0,110.38038967029796)};
                        \addlegendentry{\oplea, $\lambda=\frac{\ln n}{2}$};
                        \addplot plot [error bars/.cd, y dir=both, y explicit] coordinates
                        {(8,2.9421875)+-(0,2.172386967535653)(16,3.6193359375)+-(0,2.487270287109975)(32,3.428154296875)+-(0,2.065235147342209)(64,2.9151123046875)+-(0,1.7967420165568093)(128,8.166785888671875)+-(0,7.394710687075253)(256,11.63985336303711)+-(0,11.304182517097823)};
                        \addlegendentry{\ollga, $\lambda=\ln n$};
                        \addplot plot [error bars/.cd, y dir=both, y explicit] coordinates
                        {(8,1.88203125)+-(0,1.3576666258666696)(16,2.69390625)+-(0,2.602656167921143)(32,2.5781640625)+-(0,1.5805578106665024)(64,2.16717529296875)+-(0,1.2576939470681996)(128,7.02362060546875)+-(0,6.191703882343955)(256,9.214097900390625)+-(0,8.235461107546099)};
                        \addlegendentry{\oplea, $\lambda=\ln n$};
                        \addplot plot [error bars/.cd, y dir=both, y explicit] coordinates
                        {(8,2.9421875)+-(0,2.172386967535653)(16,2.2383984375)+-(0,1.2355714453522404)(32,2.107509765625)+-(0,0.9027205366842674)(64,1.9602294921875)+-(0,0.4797476361133488)(128,1.984171142578125)+-(0,0.430260815499792)(256,1.9704226684570312)+-(0,0.33370645814536926)(512,2.050766372680664)+-(0,0.3251223021683088)};
                        \addlegendentry{\ollga, $\lambda=\sqrt{n}$};
                        \addplot plot [error bars/.cd, y dir=both, y explicit] coordinates
                        {(8,1.88203125)+-(0,1.3576666258666696)(16,1.448046875)+-(0,0.8587667414159152)(32,1.3887890625)+-(0,0.5682796214984743)(64,1.098544921875)+-(0,0.3779829444443456)(128,1.048447265625)+-(0,0.25080198712040846)(256,0.9801536560058594)+-(0,0.2203199035690612)(512,0.9333625030517578)+-(0,0.1178163466916024)};
                        \addlegendentry{\oplea, $\lambda=\sqrt{n}$};
                        \addplot plot [error bars/.cd, y dir=both, y explicit] coordinates
                        {(8,2.33578125)+-(0,1.595321674496146)(16,1.944375)+-(0,0.8510929252910452)(32,2.06765625)+-(0,0.6304181210067659)(64,2.1286865234375)+-(0,0.4307471778575707)(128,2.2396270751953127)+-(0,0.3472940852058578)(256,2.17910400390625)+-(0,0.2411896109817295)(512,2.238797721862793)+-(0,0.16770995012756554)};
                        \addlegendentry{\ollga, $\lambda=\frac{n}{2}$};
                        \addplot plot [error bars/.cd, y dir=both, y explicit] coordinates
                        {(8,1.44453125)+-(0,0.8211301728363095)(16,1.17421875)+-(0,0.5364986584763051)(32,1.147001953125)+-(0,0.33050950843764554)(64,1.04921630859375)+-(0,0.19926095282139009)(128,1.015228271484375)+-(0,0.13996872152837936)(256,1.0048008728027344)+-(0,0.08945539022515105)(512,0.9822791290283203)+-(0,0.07214867761293452)};
                        \addlegendentry{\oplea, $\lambda=\frac{n}{2}$};

    \end{loglogaxis}

    \begin{loglogaxis}[width=0.5\linewidth, height=0.26\textheight, name=axis3,
                        at={(0, -0.29\textheight)},anchor=outer south east,
                        cycle list name=loruntimes, grid=major, log base x=2,
                        xlabel={Problem size $n$}, ylabel={Runtime $/n^2$}, title={$q = \frac{1}{6e}$}]

                        \addplot plot [error bars/.cd, y dir=both, y explicit] coordinates
                        {(8,3.1809375)+-(0,2.160169874504954)(16,5.279296875)+-(0,3.431128527883139)(32,31.834951171875)+-(0,27.992082120890693)(64,51.09688720703125)+-(0,49.54863349318307)};
                        \addplot plot [error bars/.cd, y dir=both, y explicit] coordinates
                        {(8,1.91375)+-(0,1.4708636387340601)(16,3.30421875)+-(0,2.077697271536072)(32,19.672578125)+-(0,22.053922058759255)(64,85.330810546875)+-(0,79.19181179043838)};
                        \addplot plot [error bars/.cd, y dir=both, y explicit] coordinates
                        {(8,2.2140625)+-(0,1.3022863383416299)(16,2.1056640625)+-(0,1.151041024094254)(32,2.215869140625)+-(0,1.1892033808056748)(64,2.5874560546875)+-(0,1.3414890290781023)(128,25.773662109375)+-(0,23.519267891296337)};
                        \addplot plot [error bars/.cd, y dir=both, y explicit] coordinates
                        {(8,1.445625)+-(0,0.8562028956021464)(16,1.6641796875)+-(0,0.9965851200726874)(32,1.641640625)+-(0,0.8122486127029647)(64,2.21669921875)+-(0,1.4601497023401466)(128,43.286660766601564)+-(0,43.704438038396795)};
                        \addplot plot [error bars/.cd, y dir=both, y explicit] coordinates
                        {(8,2.2140625)+-(0,1.3022863383416299)(16,1.86609375)+-(0,0.8011257033562992)(32,1.994716796875)+-(0,0.8169413394739985)(64,1.87560302734375)+-(0,0.48671505218497235)(128,2.05949951171875)+-(0,0.4583827669628903)(256,2.1604788208007815)+-(0,0.5498898440006335)(512,2.44119815826416)+-(0,0.7392532974151532)};
                        \addplot plot [error bars/.cd, y dir=both, y explicit] coordinates
                        {(8,1.445625)+-(0,0.8562028956021464)(16,1.1748046875)+-(0,0.5467380897260811)(32,1.1511328125)+-(0,0.46496525296241475)(64,1.0846142578125)+-(0,0.2569757863876345)(128,1.07876953125)+-(0,0.285179897048999)(256,1.0897732543945313)+-(0,0.3663356963683445)(512,1.1874729156494142)+-(0,0.3984180850094942)};
                        \addplot plot [error bars/.cd, y dir=both, y explicit] coordinates
                        {(8,1.845)+-(0,1.211054792706548)(16,1.9317578125)+-(0,0.9824577458349414)(32,2.152412109375)+-(0,0.5469844556862843)(64,2.18089599609375)+-(0,0.4769739686939997)(128,2.2098651123046875)+-(0,0.330157627288678)(256,2.3051805114746093)+-(0,0.2536704308289023)(512,2.3495996475219725)+-(0,0.17377809075715722)};
                        \addplot plot [error bars/.cd, y dir=both, y explicit] coordinates
                        {(8,1.15703125)+-(0,0.7896961410043659)(16,1.1351953125)+-(0,0.5125172369314572)(32,1.01501953125)+-(0,0.3321476065652093)(64,1.05227783203125)+-(0,0.23039990616941777)(128,1.052044677734375)+-(0,0.14801576722878293)(256,1.04088134765625)+-(0,0.11571887058391618)(512,1.0360332107543946)+-(0,0.07185005897804235)};

    \end{loglogaxis}
    
    \begin{loglogaxis}[width=0.5\linewidth, height=0.26\textheight, name=axis4,
                        at={(0.5\linewidth, -0.29\textheight)},anchor=outer south east,
                        cycle list name=loruntimes, grid=major, log base x=2,
                        xlabel={Problem size $n$}, 
                        title={$q = 1$}]

                        \addplot plot [error bars/.cd, y dir=both, y explicit] coordinates
                        {(8,13.94625)+-(0,13.39947842530867)(16,189.2490234375)+-(0,170.22801860352)};
                        \addplot plot [error bars/.cd, y dir=both, y explicit] coordinates
                        {(8,8.736742424242424)+-(0,9.216535420215617)(16,164.30234375)+-(0,161.7480355560017)};
                        \addplot plot [error bars/.cd, y dir=both, y explicit] coordinates
                        {(8,6.06015625)+-(0,5.745277176852932)(16,33.3734375)+-(0,41.87891135825684)(32,1322.59080078125)+-(0,1286.226692169462)};
                        \addplot plot [error bars/.cd, y dir=both, y explicit] coordinates
                        {(8,5.03953125)+-(0,4.667655627573352)(16,36.5387109375)+-(0,35.15865009209715)(32,2178.49984375)+-(0,2449.554166746444)};
                        \addplot plot [error bars/.cd, y dir=both, y explicit] coordinates
                        {(8,6.06015625)+-(0,5.745277176852932)(16,12.2804296875)+-(0,9.485627544438174)(32,106.066318359375)+-(0,109.83145299667913)(64,2389.680080566406)+-(0,2143.0401232234785)};
                        \addplot plot [error bars/.cd, y dir=both, y explicit] coordinates
                        {(8,5.03953125)+-(0,4.667655627573352)(16,9.1626953125)+-(0,7.391755627853363)(32,134.02025390625)+-(0,152.15659824362794)};
                        \addplot plot [error bars/.cd, y dir=both, y explicit] coordinates
                        {(8,5.439375)+-(0,4.9372746586236715)(16,5.9765625)+-(0,5.254492880818311)(32,5.22328125)+-(0,2.8581661066752524)(64,5.36884765625)+-(0,2.7512063973874334)};
                        \addplot plot [error bars/.cd, y dir=both, y explicit] coordinates
                        {(8,3.7078125)+-(0,3.708402423572212)(16,4.0243359375)+-(0,3.286740702503866)(32,3.518037109375)+-(0,1.7544138120972503)(64,2.8902392578125)+-(0,1.580051064533749)};

    \end{loglogaxis}

\end{tikzpicture}
\caption{Mean runtimes (number of fitness evaluation) and their standard deviation over $100$ runs of the \ollga with standard parameters $p=\frac{\lambda}{n}$, $c=\frac{1}{\lambda}$ and the \oplea  with noise rates $q \in \{0, \frac{\ln(n)}{n}, \frac{1}{6e}, 1\}$ on \leadingones with noise rates $q \in \{0, \frac{\ln(n)}{n}, \frac{1}{6e}\}$  with varying problem size $n$ normalized by $n^2$. Note that the both plots for the non-zero noise rate have a logarithmic vertical axis scale.}
\label{plot:leadingones_normalized}
\end{figure*}
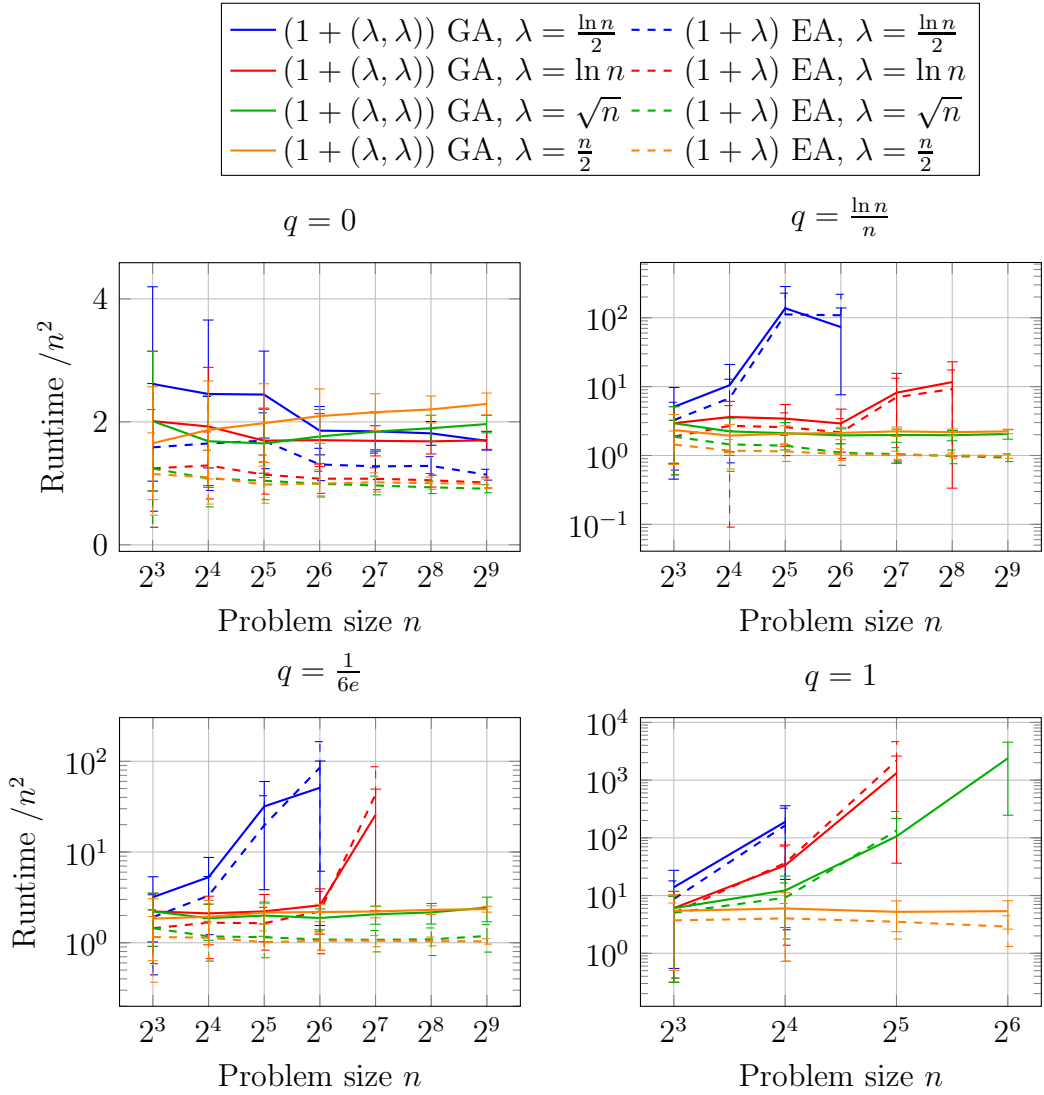

\section{Results for \jump}
\label{sec:jump}
In this section we study the performance of the \ollga, the \oplea, and the \oea on \jump functions.

We used jump functions with gap size $k=3$, since for larger values a prohibitively large number of iterations was required to find the optimum. We took the same noise rates as in Sections~\ref{sec:onemax} and~\ref{sec:leadingones}. Since there are no results on the runtime of the considered algorithms on \jump in the presence of noise, we used the following parameters. For the \ollga we used the non-standard parameters recommended for this problem in~\cite{AntipovDK22}, that is, $p=c=\sqrt{\frac{k}{n}}$. We considered two different population sizes, $\lambda = \ln(n)$, which showed a good performance on \onemax, and $\lambda=\frac{(\sqrt n)^{k-1}}{(\sqrt k) ^ k}$ also recommended for $\jump_k$ in~\cite{AntipovDK22}. We used the same population sizes for the \oplea for a fair comparison (in terms of the same order of the number of fitness evaluations made in each iteration). We run the algorithms with different noise rates on the problem sizes from $\{2^3, \dots, 2^7\}$, on which it was possible to do in a reasonable time. The results are shown in Figure~\ref{plot:jump}. This time we do not normalize the plots due to the large difference in the runtimes, but we use a logarithmic scaling for $Y$ axis.

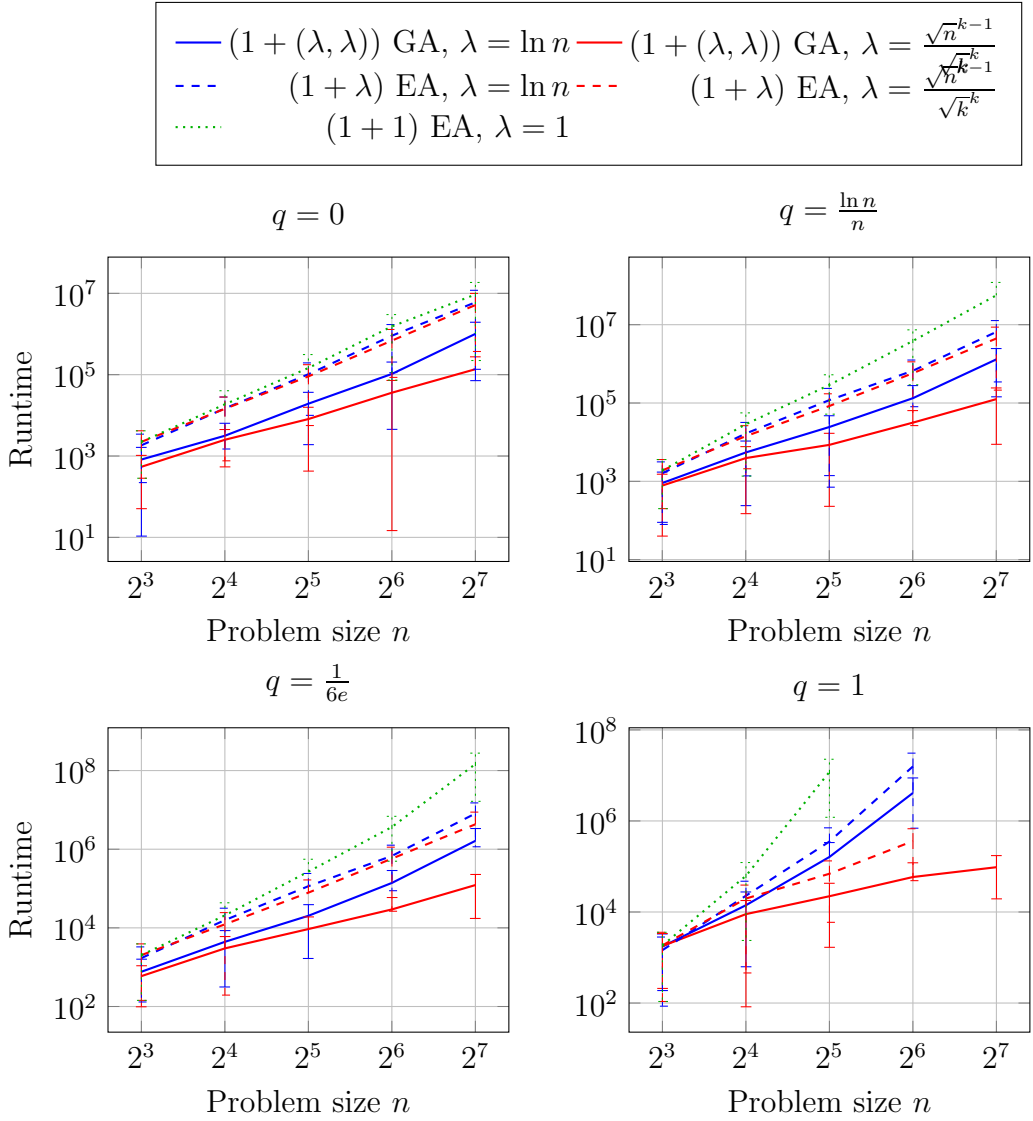
\begin{figure*}[!h]
\begin{tikzpicture}
    \begin{loglogaxis}[width=0.5\linewidth, height=0.27\textheight, name = axis1,
                        at={(0, 0)},anchor=outer south east,
                        cycle list name=jumpruntimes, grid=major, log base x=2,
                        xlabel={Problem size $n$}, ylabel={Runtime}, title={$q = 0$}]

                        \addplot plot [error bars/.cd, y dir=both, y explicit] coordinates
                        {(8,820.65)+-(0,809.9168645608017)(16,3154.35)+-(0,3229.226506069217)(32,19360.81)+-(0,17456.227626090928)(64,104434.65)+-(0,99907.5591332683)(128,1009207.8)+-(0,937830.2366078522)};
                        \addplot plot [error bars/.cd, y dir=both, y explicit] coordinates
                        {(8,543.57)+-(0,492.8226101753044)(16,2510.83)+-(0,1969.9029065159532)(32,8112.39)+-(0,7686.5849320683365)(64,36068.25)+-(0,36053.56359900502)(128,136384.64)+-(0,138388.12619134056)};
                        \addplot plot [error bars/.cd, y dir=both, y explicit] coordinates
                        {(8,1837.02)+-(0,1615.0103280165113)(16,14852.04)+-(0,13363.963575167361)(32,102468.8)+-(0,92475.50729744606)(64,910106.95)+-(0,797530.3740356524)(128,6003322.5)+-(0,5867805.381646595)};
                        \addplot plot [error bars/.cd, y dir=both, y explicit] coordinates
                        {(8,2228.6666666666665)+-(0,1944.577689085445)(16,14494.56)+-(0,13732.420343348074)(32,90816.6)+-(0,85202.0787164257)(64,689282.75)+-(0,603456.5984190806)(128,5165915.0)+-(0,4794188.85738766)};
                        \addplot plot [error bars/.cd, y dir=both, y explicit] coordinates
                        {(8,2228.6666666666665)+-(0,1944.577689085445)(16,18697.32)+-(0,21594.90199509134)(32,146129.56)+-(0,166512.10515721192)(64,1539834.22)+-(0,1466324.3322148588)(128,9366576.66)+-(0,9143572.456696387)};
                        
    \end{loglogaxis}

    \begin{loglogaxis}[width=0.5\linewidth, height=0.27\textheight, name=axis2,
                        at={(0.5\linewidth, 0)},anchor=outer south east,
                        legend style={at={(-0.1, 1.3)}, anchor=south,legend columns = 2, inner sep=7pt},
                        legend cell align={right},
                        cycle list name=jumpruntimes, grid=major, log base x=2,
                        xlabel={Problem size $n$}, 
                        title={$q = \frac{\ln n}{n}$}]

                        \addplot plot [error bars/.cd, y dir=both, y explicit] coordinates
                        {(8,904.7)+-(0,815.0974849672891)(16,5462.6)+-(0,5223.122173566306)(32,24367.98)+-(0,22968.198673809842)(64,132635.61)+-(0,146060.85915924876)(128,1301491.71)+-(0,1157528.098815742)};
                        \addlegendentry{\ollga, $\lambda=\ln n$};
                        \addplot plot [error bars/.cd, y dir=both, y explicit] coordinates
                        {(8,779.49)+-(0,739.5758851531058)(16,3921.33)+-(0,3771.915129095563)(32,8526.96)+-(0,8297.324660298644)(64,31492.75)+-(0,32521.58750257281)(128,125673.73)+-(0,116860.08075975774)};
                        \addlegendentry{\ollga, \raisebox{0pt}[0.9em]{$\lambda=\frac{\sqrt{n}^{k-1}}{\sqrt{k}^k}$}};
                        \addplot plot [error bars/.cd, y dir=both, y explicit] coordinates
                        {(8,1608.48)+-(0,1529.3076504091646)(16,16640.58)+-(0,15263.14816620739)(32,118322.76)+-(0,117614.17687431393)(64,666027.5)+-(0,585541.8240917979)(128,6549127.45)+-(0,6203720.583138214)};
                        \addlegendentry{\oplea, $\lambda=\ln n$};
                        \addplot plot [error bars/.cd, y dir=both, y explicit] coordinates
                        {(8,1900.1616161616162)+-(0,1699.531193273757)(16,14238.56)+-(0,12145.792061714214)(32,83726.3)+-(0,88282.90515535837)(64,576613.44)+-(0,550136.5431449418)(128,4442319.5)+-(0,4228977.098772793)};
                        \addlegendentry{\oplea, \raisebox{0pt}[0.9em]{$\lambda=\frac{\sqrt{n}^{k-1}}{\sqrt{k}^k}$}};
                        \addplot plot [error bars/.cd, y dir=both, y explicit] coordinates
                        {(8,1900.1616161616162)+-(0,1699.531193273757)(16,28678.61224489796)+-(0,27320.388667280677)(32,288968.62626262626)+-(0,237041.55164620426)(64,3841590.3)+-(0,3549020.3052552)(128,56327302.78)+-(0,63653939.403694235)};
                        \addlegendentry{\oea, $\lambda=1$};

    \end{loglogaxis}

    \begin{loglogaxis}[width=0.5\linewidth, height=0.27\textheight,
                        at={(0, -0.3\textheight)},anchor=outer south east,
                        cycle list name=jumpruntimes, grid=major, log base x=2,
                        xlabel={Problem size $n$}, ylabel={Runtime}, title={$q = \frac{1}{6e}$}]

                        \addplot plot [error bars/.cd, y dir=both, y explicit] coordinates
                        {(8,766.0)+-(0,830.9479526420413)(16,4418.05)+-(0,4104.0900876442765)(32,20290.06)+-(0,18624.637927122236)(64,139015.71)+-(0,148222.4667686579)(128,1635927.48)+-(0,1722014.600547408)};
                        \addplot plot [error bars/.cd, y dir=both, y explicit] coordinates
                        {(8,593.25)+-(0,494.86786872861325)(16,3006.64)+-(0,3008.0156566081896)(32,9329.84)+-(0,9656.171758745804)(64,29454.75)+-(0,29643.306937443736)(128,122657.29)+-(0,105295.77228277449)};
                        \addplot plot [error bars/.cd, y dir=both, y explicit] coordinates
                        {(8,1707.0)+-(0,1576.5821196499726)(16,15751.23)+-(0,15874.474127891608)(32,114696.88)+-(0,126773.93466302764)(64,677037.65)+-(0,588945.6871102187)(128,8105441.45)+-(0,6951934.532803812)};
                        \addplot plot [error bars/.cd, y dir=both, y explicit] coordinates
                        {(8,2028.494623655914)+-(0,1885.4592357144684)(16,12300.48)+-(0,12105.509281711364)(32,78659.49)+-(0,86840.05756153032)(64,576613.44)+-(0,550136.5431449418)(128,4310834.0)+-(0,4470955.669884125)};
                        \addplot plot [error bars/.cd, y dir=both, y explicit] coordinates
                        {(8,2028.494623655914)+-(0,1885.4592357144684)(16,20815.535353535353)+-(0,22582.81451861312)(32,269291.44)+-(0,286375.0380788214)(64,3677333.36)+-(0,3182036.016098534)(128,147041126.60869566)+-(0,130535648.15353465)};

    \end{loglogaxis}
    
    \begin{loglogaxis}[width=0.5\linewidth, height=0.27\textheight,
                        at={(0.5\linewidth, -0.3\textheight)},anchor=outer south east,
                        cycle list name=jumpruntimes, grid=major, log base x=2,
                        xlabel={Problem size $n$}, title={$q = 1$}]

                        \addplot plot [error bars/.cd, y dir=both, y explicit] coordinates
                        {(8,1741.15)+-(0,1553.041669595507)(16,14086.6)+-(0,13459.849514017607)(32,161950.6)+-(0,173311.88766959988)(64,4123429.11)+-(0,4684740.316799102)};
                        \addplot plot [error bars/.cd, y dir=both, y explicit] coordinates
                        {(8,1790.7)+-(0,1581.3211153968698)(16,8981.84)+-(0,8899.0915083732)(32,22206.08)+-(0,20530.022521994466)(64,58702.5)+-(0,61794.60408279998)(128,96407.01)+-(0,76962.2060755141)};
                        \addplot plot [error bars/.cd, y dir=both, y explicit] coordinates
                        {(8,1452.63)+-(0,1367.2488921553384)(16,22792.44)+-(0,24621.691684496418)(32,352961.6)+-(0,353065.73649222887)(64,15784854.25)+-(0,15091459.531827271)};
                        \addplot plot [error bars/.cd, y dir=both, y explicit] coordinates
                        {(8,1846.08)+-(0,1737.954381909951)(16,19738.0)+-(0,19279.843729657146)(32,69269.48)+-(0,63335.80286954291)(64,361729.68)+-(0,312929.36380291573)};
                        \addplot plot [error bars/.cd, y dir=both, y explicit] coordinates
                        {(8,1846.08)+-(0,1737.954381909951)(16,62243.08)+-(0,59873.71783039365)(32,11928811.66)+-(0,10726258.937049882)};

    \end{loglogaxis}

\end{tikzpicture}
\caption{Mean runtimes (number of fitness evaluation) and their standard deviation over $100$ runs of the \ollga with non-standard parameters recommended for \jump $p=c=\sqrt{\frac{k}{n}}$, the \oplea and the \oea on $\jump_k$ with parameter $k=3$ with varying problem size $n$.}
\label{plot:jump}
\end{figure*}

We can see from the plots that the performance of the \oea drops drastically with the growth of the noise rate, while the performance of both population-based EAs is not significantly affected by the noise for both considered population sizes (except for the large noise rate $q = 1$). 
This also implies that the relation between the runtimes of the \oplea and \ollga stays the same in the presence of noise as without, namely, the runtime of the \ollga is significantly smaller.
To support this observation, for each algorithm, each problem size and each non-zero noise rate we run statistical tests comparing them with the runtimes for $q = 0$. We use Students' t-test, which suits our study of mean values, but since this test requires the distribution of the values to be normal, we complement it with non-parametric Wilcoxon rank sum test. The obtained $p$-values are shown in Tables~\ref{tab:p-values-olea} in and~\ref{tab:p-values-ollga}. These $p$-values are more than $0.05/3 \approx 0.016$ in the most cases for the population-based algorithms except for the case when $q = 1$.  Note that we apply the Bonferroni correction and divide the standard threshold value $0.05$ by three, since we use the same samples for $q = 0$ in each of three hypotheses for each algorithm setting and each problem size.

\begin{landscape}
    \begin{table*}
        \caption{$p$-values for the experimental results in Figure~\ref{plot:jump} for the \oea and the \oplea. In the second column we denote Student's t-test by T and Wilcoxon ranksum test by W}
        \label{tab:p-values-olea}
        \begin{tabular}{|cc|ccc|ccc|ccc|}
            \hline
            \multirow{3}{*}{$n$} & \multirow{3}{*}{Test} & \multicolumn{3}{|c|}{\multirow{2}{*}{\oea}} & \multicolumn{6}{|c|}{\oplea}  \\ \cline{6-11}
            & &  & & & \multicolumn{3}{|c|}{$\lambda = \ln(n)$} & \multicolumn{3}{|c|}{$\lambda = \frac{\sqrt{n}^{k - 1}}{\sqrt{k}^k}$} \\ \cline{3-11}
            & & $q=\frac{\ln(n)}{n}$ & $q = \frac{1}{6e}$ & $q = 1$ & $q=\frac{\ln(n)}{n}$ & $q = \frac{1}{6e}$ & $q = 1$ & $q=\frac{\ln(n)}{n}$ & $q = \frac{1}{6e}$ & $q = 1$ \\ \hline
            \multirow{2}{*}{$8$} & T & $0.323$ & $0.356$ & $0.266$ & $0.308$ & $0.567$ & $0.072$ & $0.323$ & $0.356$ & $0.266$  \\
            & W & $0.499$ & $0.199$ & $0.374$ & $0.247$ & $0.599$ & $0.124$ & $0.499$ & $0.199$ & $0.374$ \\ \hline
            \multirow{2}{*}{$16$} & T & $7.83 \cdot 10^{-3}$ & $0.544$ & $1.16 \cdot 10^{-10}$ & $0.381$ & $0.667$ & $5.29 \cdot 10^{-3}$ & $0.890$ & $0.234$ & $0.029$ \\
            & W & $5.27 \cdot 10^{-3}$ & $0.494$ & $4.61 \cdot 10^{-12}$ & $0.302$ & $0.645$ & $0.018$ & $0.953$ & $0.269$ & $0.067$ \\ \hline
            \multirow{2}{*}{$32$} & T & $3.13 \cdot 10^{-6}$ & $2.80 \cdot 10^{-4}$ & $4.66 \cdot 10^{-22}$ & $0.293$ & $0.439$ & $1.03 \cdot 10^{-10}$ & $0.566$ & $0.321$ & $0.045$ \\
            & W & $1.66 \cdot 10^{-6}$ & $2.36 \cdot 10^{-4}$ & $1.35 \cdot 10^{-33}$ & $0.639$ & $0.930$ & $9.03 \cdot 10^{-13}$ & $0.318$ & $0.151$ & $0.131$  \\ \hline
            \multirow{2}{*}{$64$} & T & $1.11 \cdot 10^{-8}$ & $6.40 \cdot 10^{-9}$ & - & $0.015$ & $0.020$ & $1.02 \cdot 10^{-18}$ & $0.171$ & $0.171$ & $3.20 \cdot 10^{-6}$ \\
            & W & $3.44 \cdot 10^{-8}$ & $5.11 \cdot 10^{-8}$ & - & $0.063$ & $0.074$ & $9.05 \cdot 10^{-28}$ & $0.125$ & $0.125$ & $4.22 \cdot 10^{-5}$ \\ \hline
            \multirow{2}{*}{$128$} & T & $8.34 \cdot 10^{-12}$ & $6.64 \cdot 10^{-18}$ & - & $0.526$ & $0.023$ & - & $0.261$ & $0.196$ & - \\
            & W & $1.12 \cdot 10^{-16}$ & $9.25 \cdot 10^{-19}$ & - & $0.524$ & $0.016$ & - & $0.346$ & $0.122$ & - \\ \hline
    \end{tabular}
    \end{table*}
\end{landscape}

\begin{landscape}
    \begin{table*}
        \caption{$p$-values for the experimental results in Figure~\ref{plot:jump} for the \ollga. In the second column we denote Student's t-test by T and Wilcoxon ranksum test by W}
        \label{tab:p-values-ollga}
        \begin{tabular}{|cc|ccc|ccc|}
            \hline
            \multirow{3}{*}{$n$} & \multirow{3}{*}{Test} & \multicolumn{6}{|c|}{\ollga} \\ \cline{3-8}
            & &  \multicolumn{3}{|c|}{$\lambda = \ln(n)$} & \multicolumn{3}{|c|}{$\lambda = \frac{\sqrt{n}^{k - 1}}{\sqrt{k}^k}$} \\ \cline{3-8}
            & & $q=\frac{\ln(n)}{n}$ & $q = \frac{1}{6e}$ & $q = 1$ & $q=\frac{\ln(n)}{n}$ & $q = \frac{1}{6e}$ & $q = 1$ \\ \hline
            \multirow{2}{*}{$8$} & T & $0.468$ & $0.640$ & $4.31 \cdot 10^{-7}$ & $8.92 \cdot 10^{-3}$ & $0.480$ & $2.21 \cdot 10^{-12}$ \\
            & W & $0.494$ & $0.243$ & $4.36 \cdot 10^{-6}$ & $0.021$ & $0.341$ & $1.68 \cdot 10^{-10}$ \\ \hline
            \multirow{2}{*}{$16$} & T & $2.41 \cdot 10^{-4}$ & $0.017$ & $2.43 \cdot 10^{-13}$ & $1.15 \cdot 10^{-3}$ & $0.172$ & $2.69 \cdot 10^{-11}$ \\
            & W & $1.55 \cdot 10^{-4}$ & $0.023$ & $1.51 \cdot 10^{-14}$ & $0.021$ & $0.653$ & $2.00 \cdot 10^{-10}$ \\ \hline
            \multirow{2}{*}{$32$} & T & $0.086$ & $0.718$ & $4.20 \cdot 10^{-14}$ & $0.716$ & $0.328$ & $1.12 \cdot 10^{-9}$ \\
            & W & $0.274$ & $0.920$ & $8.18 \cdot 10^{-23}$ & $0.980$ & $0.472$ & $2.02 \cdot 10^{-9}$ \\ \hline
            \multirow{2}{*}{$64$} & T & $0.114$ & $0.056$ & $3.68 \cdot 10^{-15}$ & $0.350$ & $0.160$ & $1.90 \cdot 10^{-3}$ \\
            & W & $0.401$ & $0.227$ & $6.26 \cdot 10^{-33}$ & $0.333$ & $0.199$ & $9.60 \cdot 10^{-3}$ \\ \hline
            \multirow{2}{*}{$128$} & T & $0.052$ & $1.71 \cdot 10^{-3}$ & - & $0.557$ & $0.433$ & $0.013$ \\
            & W & $0.047$ & $2.45 \cdot 10^{-3}$ & - & $0.822$ & $0.894$ & $0.130$ \\ \hline
    \end{tabular}
    \end{table*}
\end{landscape}

\section{Conclusion}
\label{sec:conclusion}

In this work, we conducted the first experimental analysis on how robust the \ollga is to bit-wise prior noise. Our results for various noise intensities $q$ on the classic \onemax, \leadingones, and \jump benchmark show that from a logarithmic offspring population size $\lambda$ on, the \ollga is very robust to noise and can stand even constant noise rates, i.e., bit-wise noise with per-bit error rate $\Theta(1/n)$. On the \onemax and \jump problems, where this algorithm was previously shown to outperform the \oplea, it keeps this advantage also in the presence of noise. Together with the result of~\cite{AntipovDK19foga}, indicating that the \ollga on problems that are unsuitable for its main working principle can fall back to the \oplea, this work suggests that the \ollga is an interesting alternative to comparable mutation-based EAs. 

In this first work on the robustness of the \ollga to noise, we could not yet derive clear recommendations on the choice of the parameters. On the positive side, our results suggest that often simple adhoc choices like a logarithmic or a linear population size~$\lambda$ do a good job. At the same time, it is clear that the recommendations from the noise-less case cannot simply be reused (this would be $o(\log n)$ for \onemax, which appears too small in our experiments). Generally speaking, our experiments show that larger population sizes are preferable with increasing noise, but that too large population sizes can be wasteful. So determining the optimal value for this parameter is an interesting open problem. Given such functional relations can be difficult to determine via experiments, a mathematical runtime analysis might be the right tool here (where we admit that such analyses can be highly nontrivial as witnessed by the fact that a good understanding of how the \oea and the \oplea optimize \leadingones in the presence of noise was only obtained very recently~\cite{Sudholt21}). 

\section*{Acknowledgements}
    This work was supported by RFBR and CNRS, project number 20-51-15009, by a public grant as part of the Investissements d'avenir project, reference ANR-11-LABX-0056-LMH, LabEx LMH and by the Australian Research Council (ARC), grant DP190103894.


\begin{thebibliography}{DNDD{\etalchar{+}}18}

\bibitem[ABD21]{AntipovBD21gecco}
Denis Antipov, Maxim Buzdalov, and Benjamin Doerr.
\newblock Lazy parameter tuning and control: choosing all parameters randomly
  from a power-law distribution.
\newblock In {\em Genetic and Evolutionary Computation Conference, GECCO 2021},
  pages 1115--1123. {ACM}, 2021.

\bibitem[ABD22]{AntipovBD22}
Denis Antipov, Maxim Buzdalov, and Benjamin Doerr.
\newblock Fast mutation in crossover-based algorithms.
\newblock {\em Algorithmica}, 84:1724--1761, 2022.

\bibitem[AD20]{AntipovD20ppsn}
Denis Antipov and Benjamin Doerr.
\newblock Runtime analysis of a heavy-tailed ${(1+(\lambda, \lambda))}$ genetic
  algorithm on jump functions.
\newblock In {\em Parallel Problem Solving From Nature, PPSN 2020, Part~II},
  pages 545--559. Springer, 2020.

\bibitem[AD21]{AntipovD21algo}
Denis Antipov and Benjamin Doerr.
\newblock A tight runtime analysis for the ${(\mu+\lambda)}$~{EA}.
\newblock {\em Algorithmica}, 83:1054--1095, 2021.

\bibitem[ADK19]{AntipovDK19foga}
Denis Antipov, Benjamin Doerr, and Vitalii Karavaev.
\newblock A tight runtime analysis for the ${(1 + (\lambda,\lambda))}$ {GA} on
  {Leading\-Ones}.
\newblock In {\em Foundations of Genetic Algorithms, FOGA 2019}, pages
  169--182. ACM, 2019.

\bibitem[ADK22]{AntipovDK22}
Denis Antipov, Benjamin Doerr, and Vitalii Karavaev.
\newblock A rigorous runtime analysis of the ${(1 + (\lambda,\lambda))}$ {GA}
  on jump functions.
\newblock {\em Algorithmica}, 84:1573--1602, 2022.

\bibitem[BBD21]{BenbakiBD21}
Riade Benbaki, Ziyad Benomar, and Benjamin Doerr.
\newblock A rigorous runtime analysis of the 2-{MMAS}$_{\mathrm{ib}}$ on jump
  functions: ant colony optimizers can cope well with local optima.
\newblock In {\em Genetic and Evolutionary Computation Conference, GECCO 2021},
  pages 4--13. {ACM}, 2021.

\bibitem[BD17]{BuzdalovD17}
Maxim Buzdalov and Benjamin Doerr.
\newblock Runtime analysis of the ${(1+(\lambda,\lambda))}$ genetic algorithm
  on random satisfiable 3-{CNF} formulas.
\newblock In {\em Genetic and Evolutionary Computation Conference, GECCO 2017},
  pages 1343--1350. {ACM}, 2017.

\bibitem[BDGG09]{BianchiDGG09}
Leonora Bianchi, Marco Dorigo, Luca~Maria Gambardella, and Walter~J. Gutjahr.
\newblock A survey on metaheuristics for stochastic combinatorial optimization.
\newblock {\em Natural Computing}, 8:239--287, 2009.

\bibitem[BDN10]{BottcherDN10}
S\"untje B{\"o}ttcher, Benjamin Doerr, and Frank Neumann.
\newblock Optimal fixed and adaptive mutation rates for the {L}eading{O}nes
  problem.
\newblock In {\em Parallel Problem Solving from Nature, PPSN 2010}, pages
  1--10. Springer, 2010.

\bibitem[COY18]{CorusOY18fast}
Dogan Corus, Pietro~S. Oliveto, and Donya Yazdani.
\newblock Fast artificial immune systems.
\newblock In {\em Parallel Problem Solving from Nature, {PPSN} 2018, Part
  {II}}, pages 67--78. Springer, 2018.

\bibitem[DD18]{DoerrD18}
Benjamin Doerr and Carola Doerr.
\newblock Optimal static and self-adjusting parameter choices for the
  ${(1+(\lambda,\lambda))}$ genetic algorithm.
\newblock {\em Algorithmica}, 80:1658--1709, 2018.

\bibitem[DDE15]{DoerrDE15}
Benjamin Doerr, Carola Doerr, and Franziska Ebel.
\newblock From black-box complexity to designing new genetic algorithms.
\newblock {\em Theoretical Computer Science}, 567:87--104, 2015.

\bibitem[DDLS23]{DoerrDLS23}
Benjamin Doerr, Arthur Dremaux, Johannes Lutzeyer, and Aur\'elien Stumpf.
\newblock How the move acceptance hyper-heuristic copes with local optima:
  drastic differences between jumps and cliffs.
\newblock In {\em Genetic and Evolutionary Computation Conference, GECCO 2023}.
  {ACM}, 2023.
\newblock To appear.

\bibitem[DEJK23]{DoerrEJK23arxiv}
Benjamin Doerr, Aymen Echarghaoui, Mohammed Jamal, and Martin~S. Krejca.
\newblock Lasting diversity and superior runtime guarantees for the $(\mu+1)$
  genetic algorithm.
\newblock {\em CoRR}, abs/2302.12570, 2023.

\bibitem[DFK{\etalchar{+}}18]{DangFKKLOSS18}
Duc{-}Cuong Dang, Tobias Friedrich, Timo K{\"{o}}tzing, Martin~S. Krejca,
  Per~Kristian Lehre, Pietro~S. Oliveto, Dirk Sudholt, and Andrew~M. Sutton.
\newblock Escaping local optima using crossover with emergent diversity.
\newblock {\em {IEEE} Transactions on Evolutionary Computation}, 22:484--497,
  2018.

\bibitem[DHK12]{DoerrHK12ants}
Benjamin Doerr, Ashish~Ranjan Hota, and Timo K{\"o}tzing.
\newblock Ants easily solve stochastic shortest path problems.
\newblock In {\em Genetic and Evolutionary Computation Conference, GECCO 2012},
  pages 17--24. ACM, 2012.

\bibitem[DHP22]{DoerrHP22}
Benjamin Doerr, Omar~El Hadri, and Adrien Pinard.
\newblock The $(1+(\lambda,\lambda))$ global {SEMO} algorithm.
\newblock In {\em Genetic and Evolutionary Computation Conference, GECCO 2022},
  pages 520--528. {ACM}, 2022.

\bibitem[DJW02]{DrosteJW02}
Stefan Droste, Thomas Jansen, and Ingo Wegener.
\newblock On the analysis of the (1+1) evolutionary algorithm.
\newblock {\em Theoretical Computer Science}, 276:51--81, 2002.

\bibitem[DJW12]{DoerrJW12algo}
Benjamin Doerr, Daniel Johannsen, and Carola Winzen.
\newblock Multiplicative drift analysis.
\newblock {\em Algorithmica}, 64:673--697, 2012.

\bibitem[DK13]{DoerrK13cec}
Benjamin Doerr and Marvin K{\"u}nnemann.
\newblock Royal road functions and the (1~+~$\lambda$) evolutionary algorithm:
  Almost no speed-up from larger offspring populations.
\newblock In {\em Congress on Evolutionary Computation, CEC 2013}, pages
  424--431. IEEE, 2013.

\bibitem[DK15]{DoerrK15}
Benjamin Doerr and Marvin K{\"{u}}nnemann.
\newblock Optimizing linear functions with the $(1+\lambda)$ evolutionary
  algorithm---different asymptotic runtimes for different instances.
\newblock {\em Theoretical Computer Science}, 561:3--23, 2015.

\bibitem[DK20]{DoerrK20tec}
Benjamin Doerr and Martin~S. Krejca.
\newblock Significance-based estimation-of-distribution algorithms.
\newblock {\em IEEE Transactions on Evolutionary Computation}, 24:1025--1034,
  2020.

\bibitem[DLMN17]{DoerrLMN17}
Benjamin Doerr, Huu~Phuoc Le, R\'egis Makhmara, and Ta~Duy Nguyen.
\newblock Fast genetic algorithms.
\newblock In {\em Genetic and Evolutionary Computation Conference, GECCO 2017},
  pages 777--784. {ACM}, 2017.

\bibitem[DLN19]{DangLN19}
Duc{-}Cuong Dang, Per~Kristian Lehre, and Phan Trung~Hai Nguyen.
\newblock Level-based analysis of the univariate marginal distribution
  algorithm.
\newblock {\em Algorithmica}, 81:668--702, 2019.

\bibitem[DNDD{\etalchar{+}}18]{Dang-NhuDDIN18}
Rapha{\"e}l Dang-Nhu, Thibault Dardinier, Benjamin Doerr, Gautier Izacard, and
  Dorian Nogneng.
\newblock A new analysis method for evolutionary optimization of dynamic and
  noisy objective functions.
\newblock In {\em Genetic and Evolutionary Computation Conference, GECCO 2018},
  pages 1467--1474. ACM, 2018.

\bibitem[DNSW11]{DoerrNSW11}
Benjamin Doerr, Frank Neumann, Dirk Sudholt, and Carsten Witt.
\newblock Runtime analysis of the 1-{ANT} ant colony optimizer.
\newblock {\em Theoretical Computer Science}, 412:1629--1644, 2011.

\bibitem[Doe19]{Doerr19tcs}
Benjamin Doerr.
\newblock Analyzing randomized search heuristics via stochastic domination.
\newblock {\em Theoretical Computer Science}, 773:115--137, 2019.

\bibitem[Doe21]{Doerr21cgajump}
Benjamin Doerr.
\newblock The runtime of the compact genetic algorithm on {J}ump functions.
\newblock {\em Algorithmica}, 83:3059--3107, 2021.

\bibitem[Doe22]{Doerr22}
Benjamin Doerr.
\newblock Does comma selection help to cope with local optima?
\newblock {\em Algorithmica}, 84:1659--1693, 2022.

\bibitem[Dro04]{Droste04}
Stefan Droste.
\newblock Analysis of the {(1+1)} {EA} for a noisy {OneMax}.
\newblock In {\em Genetic and Evolutionary Computation Conference, GECCO 2004},
  pages 1088--1099. Springer, 2004.

\bibitem[FKK{\etalchar{+}}16]{FriedrichKKNNS16}
Tobias Friedrich, Timo K{\"{o}}tzing, Martin~S. Krejca, Samadhi Nallaperuma,
  Frank Neumann, and Martin Schirneck.
\newblock Fast building block assembly by majority vote crossover.
\newblock In {\em Genetic and Evolutionary Computation Conference, GECCO 2016},
  pages 661--668. {ACM}, 2016.

\bibitem[GK16]{GiessenK16}
Christian Gie{\ss}en and Timo K{\"{o}}tzing.
\newblock Robustness of populations in stochastic environments.
\newblock {\em Algorithmica}, 75:462--489, 2016.

\bibitem[GKS99]{GarnierKS99}
Josselin Garnier, Leila Kallel, and Marc Schoenauer.
\newblock Rigorous hitting times for binary mutations.
\newblock {\em Evolutionary Computation}, 7:173--203, 1999.

\bibitem[GS08]{GutjahrS08}
Walter~J. Gutjahr and Giovanni Sebastiani.
\newblock Runtime analysis of ant colony optimization with best-so-far
  reinforcement.
\newblock {\em Methodology and Computing in Applied Probability}, 10:409--433,
  2008.

\bibitem[HS18]{HasenohrlS18}
V{\'{a}}clav Hasen{\"{o}}hrl and Andrew~M. Sutton.
\newblock On the runtime dynamics of the compact genetic algorithm on jump
  functions.
\newblock In {\em Genetic and Evolutionary Computation Conference, {GECCO}
  2018}, pages 967--974. {ACM}, 2018.

\bibitem[JB05]{JinB05}
Yaochu Jin and J{\"{u}}rgen Branke.
\newblock Evolutionary optimization in uncertain environments -- a survey.
\newblock {\em {IEEE} Transactions on Evolutionary Computation}, 9:303--317,
  2005.

\bibitem[JJW05]{JansenJW05}
Thomas Jansen, Kenneth A.~De Jong, and Ingo Wegener.
\newblock On the choice of the offspring population size in evolutionary
  algorithms.
\newblock {\em Evolutionary Computation}, 13:413--440, 2005.

\bibitem[JW02]{JansenW02}
Thomas Jansen and Ingo Wegener.
\newblock The analysis of evolutionary algorithms -- a proof that crossover
  really can help.
\newblock {\em Algorithmica}, 34:47--66, 2002.

\bibitem[Leh10]{Lehre10}
Per~Kristian Lehre.
\newblock Negative drift in populations.
\newblock In {\em Parallel Problem Solving from Nature, PPSN 2010}, pages
  244--253. Springer, 2010.

\bibitem[Leh11]{Lehre11}
Per~Kristian Lehre.
\newblock Fitness-levels for non-elitist populations.
\newblock In {\em Genetic and Evolutionary Computation Conference, {GECCO}
  2011}, pages 2075--2082. {ACM}, 2011.

\bibitem[LOW23]{LissovoiOW23}
Andrei Lissovoi, Pietro~S. Oliveto, and John~Alasdair Warwicker.
\newblock When move acceptance selection hyper-heuristics outperform
  {M}etropolis and elitist evolutionary algorithms and when not.
\newblock {\em Artificial Intelligence}, 314:103804, 2023.

\bibitem[LW12]{LehreW12}
Per~Kristian Lehre and Carsten Witt.
\newblock Black-box search by unbiased variation.
\newblock {\em Algorithmica}, 64:623--642, 2012.

\bibitem[M{\"{u}}h92]{Muhlenbein92}
Heinz M{\"{u}}hlenbein.
\newblock How genetic algorithms really work: mutation and hillclimbing.
\newblock In {\em Parallel Problem Solving from Nature, PPSN 1992}, pages
  15--26. Elsevier, 1992.

\bibitem[NSW09]{NeumannSW09}
Frank Neumann, Dirk Sudholt, and Carsten Witt.
\newblock Analysis of different {MMAS} {ACO} algorithms on unimodal functions
  and plateaus.
\newblock {\em Swarm Intelligence}, 3:35--68, 2009.

\bibitem[NW07]{NeumannW07}
Frank Neumann and Ingo Wegener.
\newblock Randomized local search, evolutionary algorithms, and the minimum
  spanning tree problem.
\newblock {\em Theoretical Computer Science}, 378:32--40, 2007.

\bibitem[OW15]{OlivetoW15}
Pietro~S. Oliveto and Carsten Witt.
\newblock Improved time complexity analysis of the simple genetic algorithm.
\newblock {\em Theoretical Computer Science}, 605:21--41, 2015.

\bibitem[RA19]{RoweA19}
Jonathan~E. Rowe and Aishwaryaprajna.
\newblock The benefits and limitations of voting mechanisms in evolutionary
  optimisation.
\newblock In {\em Foundations of Genetic Algorithms, {FOGA} 2019}, pages
  34--42. {ACM}, 2019.

\bibitem[Rud97]{Rudolph97}
G{\"u}nter Rudolph.
\newblock {\em Convergence Properties of Evolutionary Algorithms}.
\newblock Verlag Dr.~Kov{\v a}c, 1997.

\bibitem[RW20]{RajabiW20}
Amirhossein Rajabi and Carsten Witt.
\newblock Self-adjusting evolutionary algorithms for multimodal optimization.
\newblock In {\em Genetic and Evolutionary Computation Conference, GECCO 2020},
  pages 1314--1322. {ACM}, 2020.

\bibitem[Sud13]{Sudholt13}
Dirk Sudholt.
\newblock A new method for lower bounds on the running time of evolutionary
  algorithms.
\newblock {\em {IEEE} Transactions on Evolutionary Computation}, 17:418--435,
  2013.

\bibitem[Sud21]{Sudholt21}
Dirk Sudholt.
\newblock Analysing the robustness of evolutionary algorithms to noise: refined
  runtime bounds and an example where noise is beneficial.
\newblock {\em Algorithmica}, 83:976--1011, 2021.

\bibitem[SW19]{SudholtW19}
Dirk Sudholt and Carsten Witt.
\newblock On the choice of the update strength in estimation-of-distribution
  algorithms and ant colony optimization.
\newblock {\em Algorithmica}, 81:1450--1489, 2019.

\bibitem[Wit06]{Witt06}
Carsten Witt.
\newblock Runtime analysis of the ($\mu$ + 1) {EA} on simple pseudo-{B}oolean
  functions.
\newblock {\em Evolutionary Computation}, 14:65--86, 2006.

\bibitem[Wit13]{Witt13}
Carsten Witt.
\newblock Tight bounds on the optimization time of a randomized search
  heuristic on linear functions.
\newblock {\em Combinatorics, Probability {\&} Computing}, 22:294--318, 2013.

\bibitem[Wit19]{Witt19}
Carsten Witt.
\newblock Upper bounds on the running time of the univariate marginal
  distribution algorithm on {OneMax}.
\newblock {\em Algorithmica}, 81:632--667, 2019.

\bibitem[WVHM18]{WhitleyVHM18}
Darrell Whitley, Swetha Varadarajan, Rachel Hirsch, and Anirban Mukhopadhyay.
\newblock Exploration and exploitation without mutation: solving the jump
  function in ${\Theta(n)}$ time.
\newblock In {\em Parallel Problem Solving from Nature, {PPSN} 2018, Part
  {II}}, pages 55--66. Springer, 2018.

\end{thebibliography}


\newcommand{\etalchar}[1]{$^{#1}$}

\end{document}